\RequirePackage{fix-cm}
\documentclass[twocolumn]{svjour3}          
\smartqed  
\usepackage{graphicx}

\usepackage{pifont}
\newcommand{\cmark}{\ding{51}}%
\newcommand{\xmark}{\ding{55}}%
\usepackage{booktabs}
\usepackage{natbib}
\usepackage[colorlinks,citecolor=blue]{hyperref}
\usepackage{microtype}
\usepackage{amsmath}
\usepackage{amsfonts}
\usepackage{xcolor}
\usepackage{multirow}
\usepackage{array}

\usepackage[toc,page]{appendix}

\begin{document}

\title{What comprises a good talking-head video generation?: \textit{A Survey and Benchmark}} 



\author{Lele Chen$^*$      \and
        Guofeng Cui$^*$  \and 
        Ziyi Kou$^*$  \and
        Haitian Zheng \and
        Chenliang Xu
}

\authorrunning{Lele Chen {\it et al.}} 

\institute{
$^*:$ Equal contribution  \at
        \and
Lele Chen \at
              \email{lchen63@ur.rochester.edu}           
           \and
           Guofeng Cui \at
              \email{gcui2@ur.rochester.edu}
         \and
           Ziyi Kou \at
              \email{zkou2@ur.rochester.edu}
        \and
           Haitian Zheng \at
              \email{hzheng15@ur.rochester.edu}
         \and
           Chenliang Xu \at
              \email{chenliang.xu@rochester.edu}
}

\date{Received: date / Accepted: date}
\maketitle

\begin{abstract}
Over the years, performance evaluation has become essential in computer vision, enabling tangible progress in many sub-fields. While talking-head video generation has become an emerging research topic, existing evaluations on this topic present many limitations. For example, most approaches use human subjects (e.g., via Amazon MTurk) to evaluate their research claims directly. This subjective evaluation is cumbersome, unreproducible, and may impend the evolution of new research. In this work, we present a carefully-designed benchmark for evaluating talking-head video generation with standardized dataset pre-processing strategies. As for evaluation, we either propose new metrics or select the most appropriate ones to evaluate results in what we consider as desired properties for a good talking-head video, namely, identity preserving, lip synchronization, high video quality, and natural-spontaneous motion. 
By conducting a thoughtful analysis across several state-of-the-art talking-head generation approaches, we aim to uncover the merits and drawbacks of current methods and point out promising directions for future work. All the evaluation code is available at: \href{https://github.com/lelechen63/talking-head-generation-survey}{https://github.com/lelechen63/talking-head-generation-survey}.
\end{abstract}

\section{Introduction}
\label{sec:intro}


Given one (or a few) facial image(s) and a driving source (e.g., a piece of audio speech or a sequence of facial landmarks), the task of talking-head video generation is to synthesize a realistic-looking, animated talking-head video that corresponds to the driving source. Solving this task is crucial to enabling a wide range of practical applications, such as re-dubbing videos with other languages, telepresence for video-conferencing or role-playing video games, bandwidth-limited video transformation, and virtual anchors. Another potential application is enhancing speech comprehension while preserving privacy or assistive devices for hearing impaired people.  Meanwhile, it can benefit the research of adversarial attacks in security and provide more training samples for supervised learning approaches. However, studying such a video synthesis problem is known to be challenging for the following three reasons. First, the deformation of a talking-head consists of individual's intrinsic subject traits, extrinsic camera positions, head movements, and facial expressions, which are highly convoluted. 
This complexity stems not only from modeling face regions but also from modeling the
head motion and background. Second, explicitly exploiting the visual information contained in the reference video remains unsolved. 
Lastly, it is challenging because the problem of subtle artifacts and perceptual identity changes that people are sensitive to in a synthesized video is hard to avoid in learning-based methods. 


The graphics-based talking head generation approaches mainly focus on subject-dependent video-editing, which requires a full original video sequence as input (Bregler et al.~\citeyear{bregler1997video}; Chang and Ezzat~\citeyear{chang2005transferable};  Liu and Ostermann et al.~\citeyear{liu2011realistic}; Garrido et al.~\citeyear{garrido2015vdub}; Suwajanakorn et al.~\citeyear{suwajanakorn2017synthesizing}; Fried et al.~\citeyear{fried2019text}). 
For instance, Suwajanakorn et al.~\citeyearpar{suwajanakorn2017synthesizing} generate a lip region image from an audio signal, and compose it with a retrieved frame from a large video corpus of the target person, to produce the final video frame. While the proposed method can synthesize fairly photo-realistic videos, it requires a large amount of video footage of the target person to compose the final video. 
Recently, several audio-driven face generation works are proposed to synthesize identity-independent facial animation with fixed head pose (Chung et al.~\citeyear{jamaludin2019you}; Pumarola et al.~\citeyear{ganimation}; Chen et al.~\citeyear{chen2018lip}; Song et al.~\citeyear{song2018talking}; Vougioukas et al.~\citeyear{vougioukas2019realistic}; Chen et al.~\citeyear{chen2019hierarchical}; Zhou et al.~\citeyear{zhou2019talking}). 
For example, Vougioukas et al.~\citeyearpar{vougioukas2019realistic} propose a temporal generative adversarial network (GAN), capable of generating a talking-head video with natural facial expressions from an audio signal. 
The generated video
conveys a plethora of information not only about the phonemic information but also about the emotional expression. 
To consider head motion modeling, landmark-driven methods (Wang et al.~\citeyear{wang2019few}; Zakharov et al.~\citeyear{zakharov2019few}; Gu et al.~\citeyear{gu2019flnet}) are introduced that control the head motion and facial expression with facial landmarks. 
Zakharov et al.~\citeyearpar{zakharov2019few} even show that their approach is able to synthesize realistic talking-head video of portrait paintings, with control of the head movement. 
In this paper, we focus on surveying and evaluating identity-independent talking-head generation methods.


A sizable volume of follow-up papers have been published since the introduction of 
identity-independent talking-head generation task (Chung et al.~\citeyear{jamaludin2019you}). 
A large number of talking-head generative model variants have been
introduced, including few-shot models, recurrent models, and 3D graphics combined models. 
While there has been substantial progress in terms of synthesized video quality, relatively less effort has been spent in evaluating talking-head methods, and grounded ways to quantitatively assess these videos are still missing. 
Several evaluation measures have surfaced with the flourish of talking-head generation models. Some of the metrics attempt to quantitatively evaluate synthesized images, while others emphasize qualitative ways such as user studies or analyzing internals of models. 
Both of these evaluations present strengths and limitations. 
For example, qualitative evaluation often resorts to manual inspection of the visual fidelity of generated images. 
People may think that fooling a person (e.g., via Amazon MTurk) in distinguishing generated video frames from ground truth can be the ultimate test. 
However, such Turing tests tend to
be biased towards the models that concentrate on limited sections of the data (e.g., memorizing or over-fitting; low diversity) and neglect the overall distributional characteristics, which 
are essential for unsupervised learning. 
Meanwhile, such evaluation is time-consuming and possibly misleading. Quantitative metrics, while being less subjective, may not be able to clearly specify in which scenarios the scores are meaningful and in which other scenarios prone to misinterpretations, since they may not directly correspond to how humans perceive and judge generated video frames.



The goal of this paper is to comprehensively examine existing literature on quantitative measures of talking-head generative models, and help researchers assess them 
objectively. Before delving into the survey, we list four desired properties that
a good synthesized talking-head video should fulfill: preserving subjects' original identity, maintaining lip-synced at a semantic-level, keeping high visual quality, and containing spontaneous motions. 
These properties can serve as meta measures to evaluate and compare different talking-head generation approaches. 
By assessing these properties on 
quantitative aspects, we hope to answer the following questions: 1) what are the strengths and limitations of current evaluation metrics? 2) which metrics should be preferred accordingly, or is there any better evaluation metric that can be introduced to evaluate talking-head video generation approaches? 3) are the metrics robust to different testing protocols? 

While some existing metrics are shown to be effective image-level visual quality evaluation, there are some other issues, such as the variety of probability criteria and the lack of perceptually meaningful video-level measures, have made evaluating the talking-head video generative models notoriously tricky. 
In this paper, we mainly discuss and assess talking-head video generative approaches by either designing or choosing evaluation metrics concerning the four desiderata: 
\begin{enumerate}
    \item \textbf{Identity Preserving.} We compare two existing identity-preserving evaluation metrics by visualizing the decision boundaries of inter-class discrepancy ability, and select cosine similarity between embedding vectors of ArcFace (Deng et al.~\citeyear{deng2019arcface}) to measure identity mismatch.
    
    \item \textbf{Visual Quality.} We use SSIM and FID to evaluate visual quality at an image-level since they are sensitive to image distortions and transformations. And we use CPBD to judge the sharpness of the synthesized video.
    
    \item \textbf{Semantic-level Lip Synchronization.} While some methods can generate realistically looking videos, the generated lip movements usually present less expressive and discriminative semantic cues, which can not convey the audio information. 
    To address this semantic lip-synchronization ability, we critically discuss existing lip-sync evaluation methods and introduce a new lip-sync metric---lipreading Similarity Distance (LRSD), which evaluates the lip movement synchronization in semantic perspective. 
    The experimental results demonstrate that our LRSD score agrees with human perceptual judgments and human rankings of videos.
    
    \item \textbf{Natural-spontaneous Motion.} Video generative models have well-known limitations, including a tendency towards limited diversity in generated video samples. In order to investigate intra-video diversity, we evaluate the spontaneous motions emitted in synthesized videos, including emotional expression, blinks, and head movements. Meanwhile, we design a new evaluation metric---Emotion Similarity Distance (ESD) to evaluate the facial emotional expression distance between the synthesized video and the ground truth. To quantitatively evaluate the subconscious blinks in a talking-head video, we introduce a learning-based metric---Blink Similarity Distance (BSD)to evaluate the quality of the blink motion in the eye region of a synthesized video.
\end{enumerate}

In addition to evaluation metrics, we conduct thoughtful experiments to evaluate state-of-the-art talking-head generation approaches under different protocols. 
There are several interesting findings that have not been emphasized in previous works: Most of the current deep generative models do not perform well when the head pose of reference frames and target frame are different; When we evaluate a talking-head generation method with the quantitative metrics (e.g., SSIM, FID), we should consider the distributions of head pose, head motion of the testing set; All the selected methods can not synthesize accurate lip movements for words (e.g., `JOB', `IMAGINE', and `HAPPENS'), and this could be a good direction for future researches to improve the semantic-level visual quality.

This work attempts to track recent advances and provides an in-depth look at identity-independent talking-head synthesis methods. 
We critically evaluate the quality of the synthesized video with perceptual metrics, pointing out what comprises a good talking-head video generative model. The contributions of this paper are mainly three-fold: 

\begin{itemize}
    \item To address the lack of perceptually meaningful video similarity measures, we introduce three new metrics (LRSD, ESD, and BSD) to assess the synthesized video quality at a video-level, like how humans perceive and judge videos. And all the evaluation metrics will be publicly available to facilitate the community.
    \item By carefully conducting a series of experiments, we have some interesting findings. 
    For example, current learning-based talking-head generation approaches are suffering from the massive head motions in the video. We hope that those findings can open up objective directions for future works.
    \item We build a code repository, that contains the techniques we surveyed for talking-head generation, including image matting, few-shot generator, attention-based embedding, and 3D graphics module. The code repository not only provides a uniformed structure, making it easier to benchmark different talking-head video generation, but also may benefit other video dynamics generation tasks like video re-targeting (Chan et al.~\citeyear{chan2019everybody}), pose-guided human image generation (Siarohin et al.~\citeyear{siarohin2019appearance}), and video prediction (Liu et al.~\citeyear{liu2018future}; Minderer et al.~\citeyear{minderer2019unsupervised}). 
    
\end{itemize}

Ultimately, we hope that this paper can draw a clearer picture of the current panorama in talking-head generation and its evaluation, helping the researchers objectively assess and improve talking-head generation methods.

\section{What comprises a good talking-head video generation?}
\label{sec:generation}
\begin{figure}
  \includegraphics[width=0.98 \linewidth]{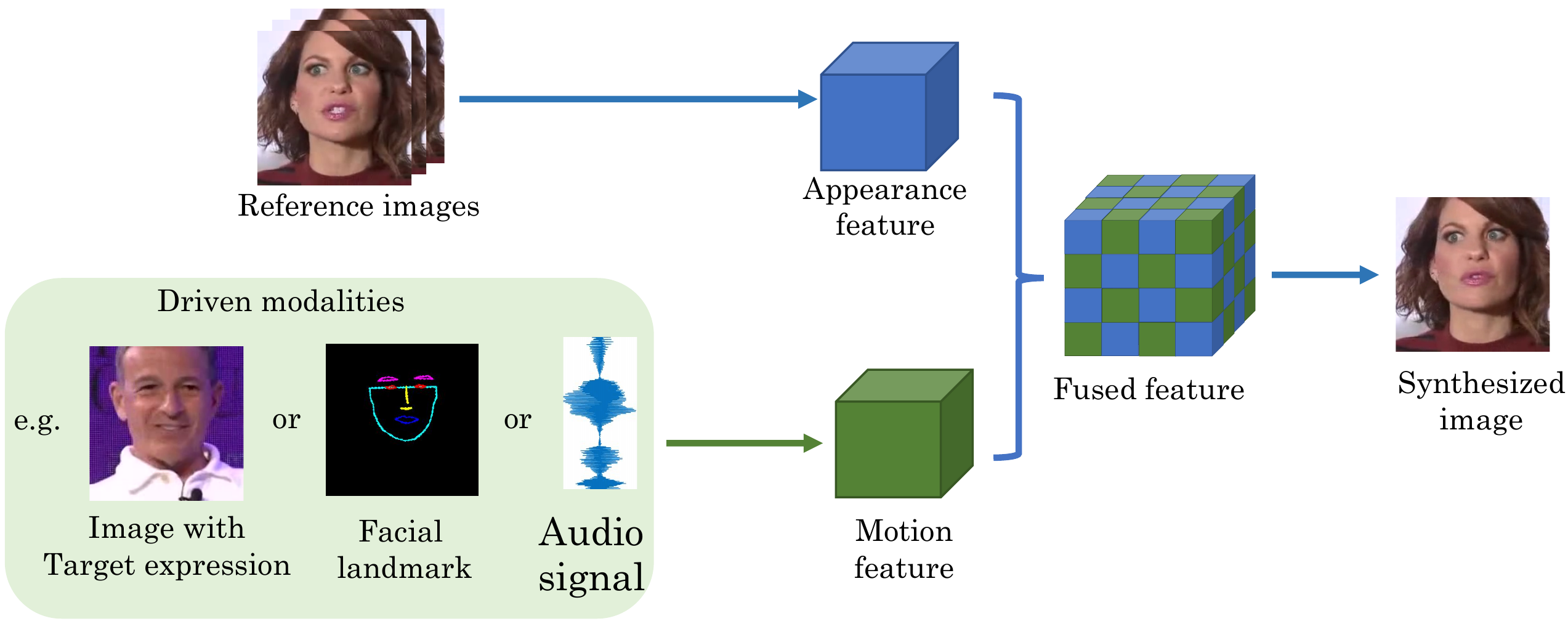}
\caption{ The general framework of talking-head generation methods.}
\label{fig:overall}       
\end{figure}

Talking-head video generation methods leverage the motion information from the driven-modality (e.g., an audio signal or facial landmarks) as well as the appearance information from 
reference images to generate new images that convey the driven-modality (Fig.~\ref{fig:overall}). In this section, we will discuss the technical contributions proposed by recent methods concerning each property listed in Sect.~\ref{sec:intro}.

\subsection{Identity Preserving}
\begin{figure}
  \includegraphics[width=0.98 \linewidth]{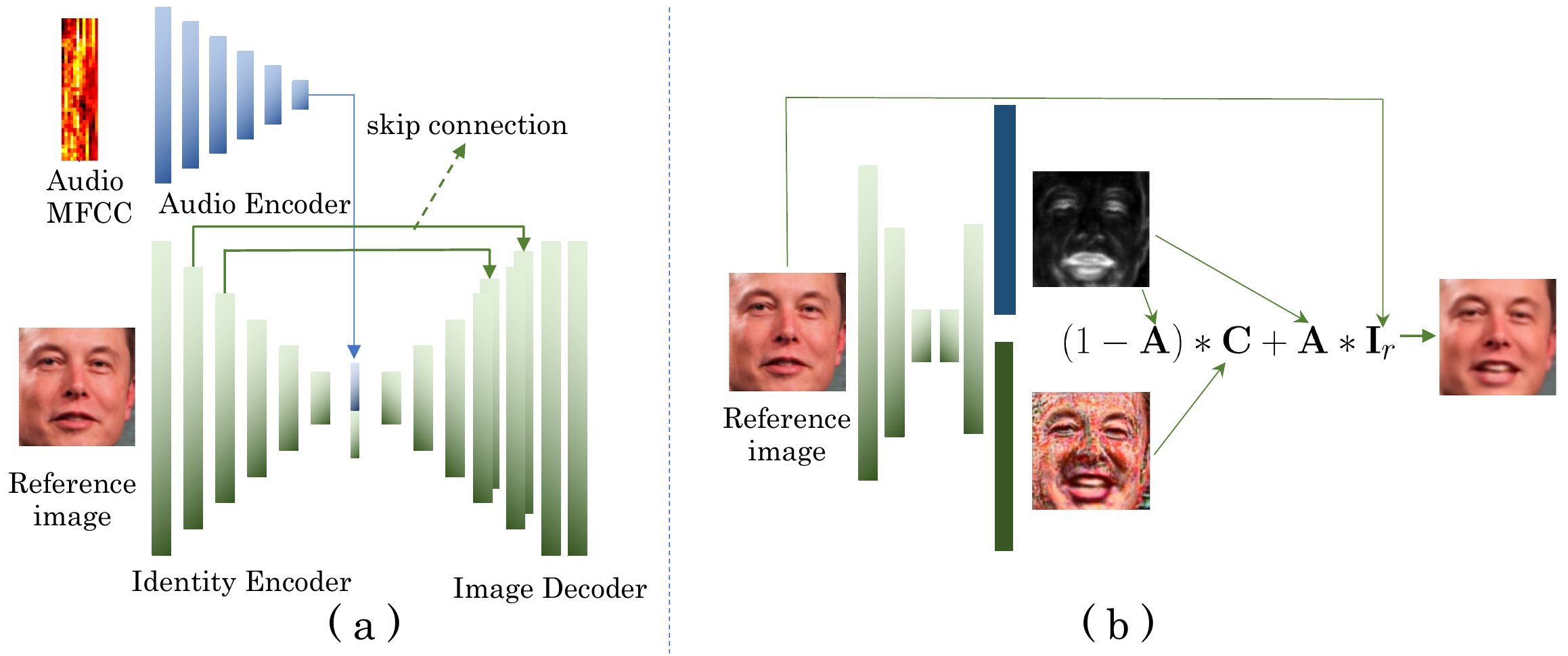}
\caption{The network illustration of skip connection and image matting function. (a) shows the detailed network structure of Jamaludin et al.~\citeyearpar{jamaludin2019you} and the skip connections design. (b) illustrates the image matting function, where $\mathbf{A}$ is the attention map obtained by applying Sigmoid activation function and $\mathbf{C}$ is the color mask obtained by applying Tanh activation function.}
\label{fig:skip}       
\end{figure}

People are sensitive to any perceptual identity changes in a synthesized video, is hard to avoid in the deep generative model. 
The reason is that the spatial identity information may not be preserved perfectly after deep convolution layers (e.g., encoding and fusion network). 
To address this problem, Jamaludin et al.~\citeyearpar{jamaludin2019you} apply skip connections (Fig.~\ref{fig:skip}a) to enrich the appearance information from the reference image. 
The skip connections offer a high-way between the synthesized image feature and the reference image feature, which can mitigate the identity information loss during the identity encoding and image decoding stage.
However, this skip connection structure is sensitive to which layer that people apply to it. 
If the skip connection is applied too early or too late, it may decrease network performance. 

Different reference images may carry different appearance features, and have different degrees of relevance to target images. 
Besides enriching identity feature at the last several layers of an image decoder, some existing works (Jamaludin et al.~\citeyear{jamaludin2019you}; Wiles et al.~\citeyear{x2face}) utilize multiple reference images to mitigate the identity loss problem.
For example, instead of using a single reference image to specify the unique identity, Jamaludin et al.~\citeyearpar{jamaludin2019you} concatenate multiple distinct images in channel-wise as identity references. The multiple reference images serve to enhance the global appearance feature and to reduce the minor variations caused by non-audio-correlated movement.
However, the concatenation operation requires aligned face images as input as it ignores the head movements. To generate talking-head with head movements, Wiles et al.~\citeyearpar{x2face} propose an embedding network to aggregate a shared representation across different reference frames with different poses and expressions. A driving network is then designed to sample pixels from the embedded common representation to produce a generated frame.
The proposed embedding network provides a novel mapping mechanism and a memory-network-like structure for extracting appearance feature. 

Very recently, some remarkable realistic results have been demonstrated by the few-shot embedding structure (Zakharov et al.~\citeyear{zakharov2019few}; Wang et al.~\citeyear{wang2019few}; Liu et al.~\citeyear{liu2019few}; Yoo et al.~\citeyear{yoo2019few}). 
For instance, Zakharov et al.~\citeyearpar{zakharov2019few} embed the reference images to predict parameters of adaptive instance normalization (AdaIN) layers in the generator. 
Similarly, Wang et al.~\citeyearpar{wang2019few} predict 
network parameters (the scale and bias map of the spatial-adaptive normalization block proposed by Park et al.~\citeyear{spade}) of the generator. Comparing with non-few-shot methods, the part of the network parameters of the generator converges much faster to the state that generates realistic and personalized images using few-shot methods.

\subsection{Visual Quality}
A major difference between image generation and video generation is the smooth transition between adjacent frames, since people are also sensitive to any pixel jittering (e.g., temporal discontinuities and subtle artifacts) in a video. Recent works (Song et al.~\citeyear{song2018talking}; vougioukas et al.~\citeyear{vougioukas2019realistic}; Chen et al.~\citeyear{chen2019hierarchical}; Wang et al.~\citeyear{wang2019few}; Zhou et al.~\citeyear{zhou2019talking}) model the temporal-dependency to achieve smoother facial transition across time. 
Specifically, Song et al.~\citeyearpar{song2018talking} propose a recurrent-generator that considers the temporal dependencies at the generation stage and a spatial-temporal discriminator that judges the synthesized video at a
video-level during discrimination stage. 
Similarly, Vougioukas et al.~\citeyearpar{vougioukas2019realistic} propose a sequence-discriminator that consists of spatial-temporal convolutions and GRUs to extract transient features and to determine if a sequence is real or not. 

In addition to temporal modeling, another skill that can improve temporal coherence is image matting function, which has been explored in Wang et al.~\citeyearpar{wang2018high}, Pumarola et al.~\citeyearpar{ganimation}, Vondrick et al.~\citeyearpar{vondrick2016generating}, Wang et al.~\citeyearpar{wang2019few}, and Chen et al.~\citeyearpar{chen2019hierarchical}. For instance, Pumarola et al.~\citeyearpar{ganimation} compute the final output image by:
\begin{equation}
\centering
\begin{aligned}
   \hat{\mathbf{I}} = (1 - \mathbf{A}) * \mathbf{C} + \mathbf{A} * \mathbf{I}_r \enspace,
\end{aligned}
\label{eq:matting}    
\end{equation}
where $\mathbf{I}_r$, $\mathbf{A}$ and $\mathbf{C}$ are the input reference image, attention map and color mask, respectively. 
The attention map $\mathbf{A}$ indicates to what extent each pixel of $\mathbf{I}_r$ contributes to the output image $\hat{\mathbf{I}}$. By applying this image matting function (Fig.~\ref{fig:skip}b) in the generator, the reused pixels from the reference image can partially stabilize the video quality. 
However, the attention mechanism may not perform well and even introduce artifacts if there is misalignment between the reference frame $\mathbf{I}_r$ and the target frame ${\mathbf{I}}$ due to large deformation caused by head movements. 
We attribute this problem to the poor composition ability of the linear image matting function when solving the misalignment between $\mathbf{I}_r$ and ${\mathbf{I}}$. 

To minimize the misalignment, rather than using the reference frame ${\mathbf{I}_r}$, Wang et al.~\citeyearpar{NIPS2018_7391} propose a sequential generative model, adopting flow-based warping on a previous synthesized frame $\hat{\mathbf{I}}_{t-1}$ to align it with the target image ${\mathbf{I}_t}$. Thus, the image matting function can be reformulated as:
\begin{equation}
\centering
\begin{aligned}
   \hat{\mathbf{I}_t} = (1 - \mathbf{A}) * \mathbf{C} + \mathbf{A} * \tilde{\mathbf{w}} (\hat{\mathbf{I}}_{t-1})  \enspace,
\end{aligned}
\label{eq:matting2}    
\end{equation}
where $\tilde{\mathbf{w}}$ is the estimated optic flow from $\hat{\mathbf{I}}_{t-1}$ to ${\mathbf{I}_t}$. By applying the image matting function on the synthesized, previous frame, they can mitigate the misalignment problem. 
However, the estimated optic flow may not able to handle 
small misalignment in face areas, which is the main factor that causes jittery artifacts between frames. Meanwhile, the warping operation may introduce extra artifacts when the estimated optic flow is not accurate. 

On the other hand, 3D graphics modeling has been introduced in GAN-based methods due to its stability (Fried et al.~\citeyear{fried2019text}; Kim et al.~\citeyear{kim2018deep}; Yi et al.~\citeyear{yi2020audio}). 
The prior texture information from 3D graphics modeling can ease the training of generator and improve the temporal coherence.

\subsection{Lip Synchronization}
Another challenge of talking-head generation is to maintain the synchronization between visual dynamics (e.g., facial movement, lip movement) and the driven modality (e.g., audio signal, and landmark) since people are sensitive to the slight misalignment between facial movements and speech audio. 
Chen et al.~\citeyearpar{chen2018lip} propose a correlation loss between the derivative of visual feature (optical flow) and the derivative of the audio feature to address the lip-sync problem. 
However, this method requires a fixed-length of audio input and can only generate fixed length image frames. 
In order to generate images sequentially, Song et al.~\citeyearpar{song2018talking} propose a conditional recurrent adversarial network to incorporate both image and audio in the recurrent unit to achieve temporal dependency in the generated video on both facial and lip movements, and further generate lip-synced video frames. 
Furthermore, they design a lipreading discriminator to boost the accuracy of lip synchronization. Similarly, Vougioukas et al.~\citeyearpar{vougioukas2019realistic} propose a sync-discriminator on fixed-length snippets of the original video and audio to determine whether they are in or out of sync. 
The proposed discriminator computes the embedding for audio and video using a two-stream architecture and then calculate the Euclidean distance between two embeddings.

\subsection{Natural-spontaneous Motions}
People naturally emit spontaneous motions such as head movements and emotional expressions when they speak, which contain nonverbal information that helps the audience comprehend the speech content (Cassell et al~\citeyear{cassel1999}; Ginosar et al~\citeyear{ginosar2019learning}). 
Although speech contains necessary information for generating lip movements, it can hardly be used to produce natural-spontaneous motions. 
Some works (Fan et al.~\citeyear{fan2015photo}; Jamaludin et al.~\citeyear{jamaludin2019you}; Song et al.~\citeyear{song2018talking}; Chen et al.~\citeyear{chen2019hierarchical}; Zhou et al.~\citeyear{zhou2019talking}) ignore the modeling of spontaneous expressions, resulting in faces that are mostly static except for the mouth region.
To model emotional expressions, Jia et al.~\citeyearpar{jia2014head} use neural networks to learn a mapping from emotional state (pleasure-displeasure, arousal-nonarousal, and dominance-submissiveness) parameters to facial expressions. Karras et al.~\citeyearpar{karras2017audio} propose a network to synthesize 3D vertex by inferring the information from the audio signal and emotional state.
Vougioukas et al.~\citeyearpar{vougioukas2019realistic} propose a noise generator capable of producing noise that is temporally coherent through a single-layer GRU. This latent representation introduces randomness in the face synthesis process and helps with the generation of blinks and brow movements. Some works (Yi et al.~\citeyear{yi2020audio}; Thies et al.~\citeyear{thies2016face2face}; Kim et al.~\citeyear{kim2018deep}; Averbuch et al.~\citeyear{averbuch2017bringing}; Zhang et al.~\citeyear{zhang2019one}) take image frames that contain the target motion as dense mapping to guide the video generation, producing video frames with convoluted head motion and facial expressions. 

\begin{figure*}
  \includegraphics[width=0.98 \linewidth]{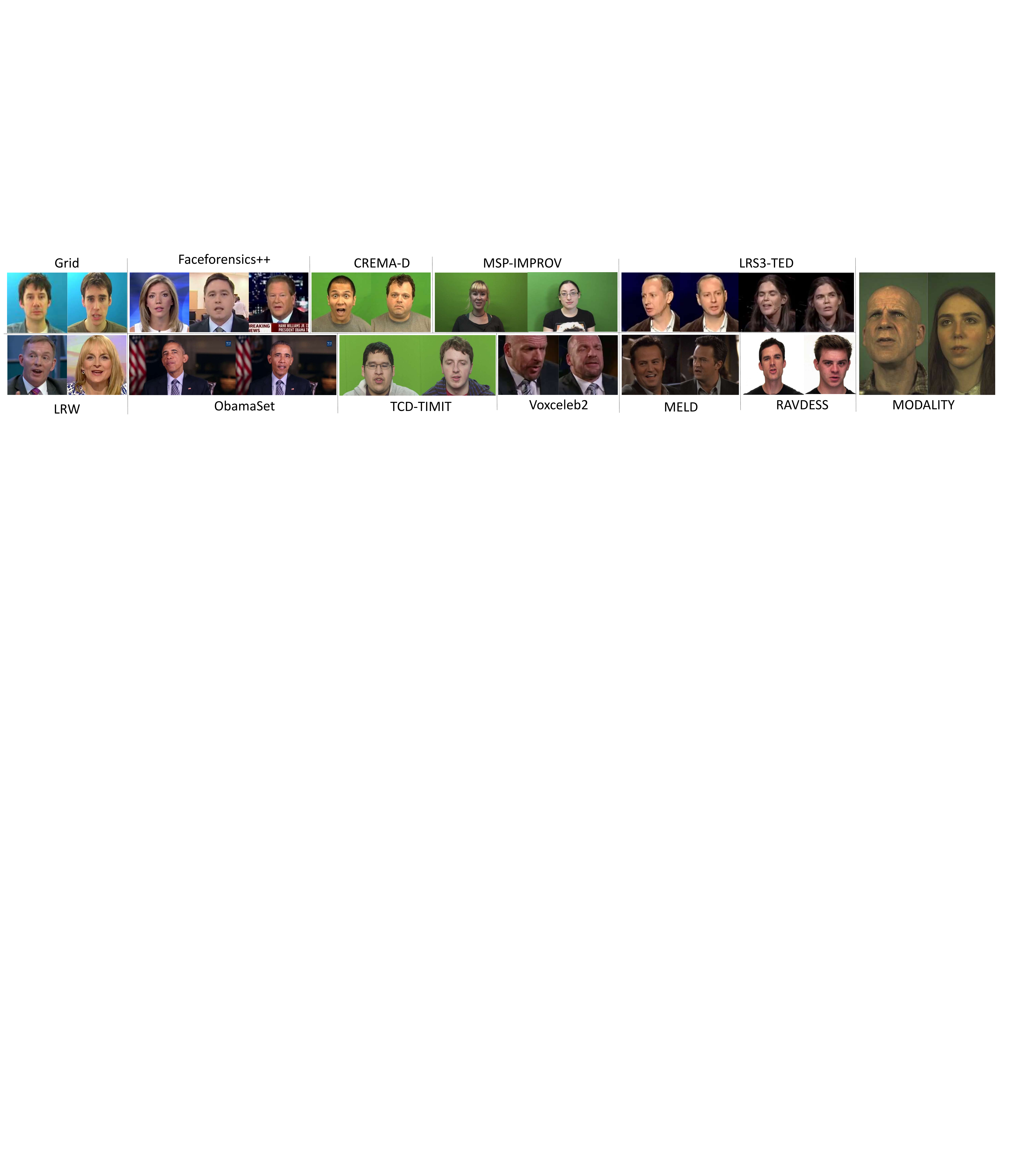}
\caption{Example images of different datasets. For each dataset of Table~\ref{Table:dataset}, several frames of video are sampled and represented.}
\label{fig:data_plot}       
\end{figure*}

\section{Review of Audio-Visual Corpora}
\label{sec:dataset}

The recent large-scale audio-visual datasets played a crucial role in the success of talking-head video generation. On the one hand, the increasingly enriched datasets capture the diversity of visual contents in lighting conditions, identities, poses, video qualities, and phrases, enabling the training of robust talking-head models for realistic scenes. On the other hand, the audio, as well as other annotated attributes of the datasets provides meaningful ways for examing and comparing the performance of different algorithms.
In this section, we summarize the attributes of the recently released speech-related audio-visual datasets, ranging from lab-controlled to in-the-wild environment data (Tab.~\ref{Table:dataset} and Fig.~\ref{fig:data_plot}) and select some representative ones (Tab.~\ref{Table:selected_data}) as the benchmark datasets for Sect.~\ref{sec:evaluation}.


\begin{table*}[t]
	\centering
	\caption{The statistics of some existing datasets. }
	\begin{tabular}{ lccccccccc}
		\toprule
		\hline
Dataset Name& Hours & Subj. & Sent.  & Vocab. & Aign.  & Move.  & Ext. Poses & Emo.  & Env. \\
GRID (Cooke et al.~\citeyear{grid})             & 27.5  & 33    & 33k    & 51     & \cmark & \xmark & \xmark & \xmark & Lab  \\
TCD-TIMIT (Harte and Gillen~\citeyear{tcd})        & 11.1  & 62    & 6.9k   & N/A    & \cmark & \xmark & \xmark & \xmark & Lab  \\
MODALITY (Czyzewski et al.~\citeyear{czyzewski2017audio})  & 31  & 35    & 5.8k    & 182     & \cmark & \xmark & \xmark & \xmark & Lab  \\
LRW (Chung and Zisserman~\citeyear{chung2016lip} )         & 173   & 1k+ & 539k   & 500    & \xmark & \xmark & \cmark & \xmark & Wild \\
CREMA-D (Cao et al.~\citeyear{crema})        & 11.1  & 91    & 12     & N/A    & \cmark & \cmark & \xmark & \cmark & Lab  \\
RAVDESS (Livingstone et al.~\citeyear{ryerson})   & 7     & 24    & 2      & 8      & \cmark & \cmark & \xmark & \cmark & Lab  \\
MSP-IMPROV (Busso et al.~\citeyear{busso2016msp})  & 18    & 12    & 652    & N/A    & \cmark 
& \cmark & \xmark & \cmark & Lab  \\
Faceforensics++ (Rossler et al.~\citeyear{facefor})& 5.7   & 1k  & 1k+  & N/A    & \xmark & \cmark & \xmark & \xmark & Wild \\
ObamaSet (Suwajanakorn et al.~\citeyear{suwajanakorn2017synthesizing})     & 14    & 1     & N/A    & N/A    & \xmark & \cmark & \xmark & \xmark & Wild \\
VoxCeleb1 (Nagrani et al.~\citeyear{voxceleb1})   & 352   & 1.2k  & 153.5k & N/A    & \xmark & \cmark & \cmark & \xmark & Wild \\
VoxCeleb2 (Chung et al.~\citeyear{voxceleb2})   & 2.4k  & 6.1k  & 1.1m   & N/A    & \xmark & \cmark & \cmark & \xmark & Wild \\
LRS2-BBC (Afouras et al.~\citeyear{afouras2018deep})          & 224.5   & 500+ & 140k+  & 59k    & \cmark & \cmark & \cmark & \xmark & Wild \\
LRS3-TED (Afouras et al.~\citeyear{lrs3})          & 438   & 5k+ & 152k+  & N/A    & \cmark & \cmark & \cmark & \xmark & Wild \\
MELD (Poria et al.~\citeyear{MELD2018})         & 13.7  & 407   & 13.7k  & 17k    & \cmark & \cmark & \cmark & \cmark & Wild \\
		\hline
        \bottomrule
	\end{tabular}
	\label{Table:dataset}
\end{table*}

\begin{table*}
	\centering
	\caption{The benchmark datasets under different scenarios. }
	\begin{tabular}{ c|cccc}
		\toprule
		\hline
Scenarios with & Fixed head pose & Spontaneous motion & Apparent head pose    & Specific person \\
Dataset  & GRID \& LRW     & CREMA-D             & VoxCeleb2 \& LRS3-TED & ObamaSet        

	\end{tabular}
	\label{Table:selected_data}
\end{table*}
\subsection{Videos Without Head Movement} 
\label{subsec:no_move}
Almost- but not entirely-natural head movement is often perceived as particularly creepy, an effect known as the uncanny valley ({\"O}hman and Salvi~\citeyear{ohman1999using}).
There are various visual dynamics in talking-head videos (e.g., camera angles, head movements) that are not relevant to and hence cannot be inferred from speech audios. 
Many works (Jamaludin et al.~\citeyear{jamaludin2019you}; Song et al.~\citeyear{song2018talking}; Zhou et al.~\citeyear{zhou2019talking}; Chen et al.~\citeyear{chen2019hierarchical}; Vougioukas et al.~\citeyear{vougioukas2019realistic}) are focusing on generating videos without any head movements. 
We list some popular datasets with fixed head poses and select the benchmark datasets to represent datasets without head movement.

In the GRID dataset (Cooke et al.~\citeyear{grid}), there are 33 speakers, front-facing the camera and each uttering 1000 short phrases, containing six words randomly chosen from a limited dictionary (51 words). 
All the sentences are spoken with neutral emotion without any noticeable head movements. 
To enrich the linguistic information, the TCD-TIMID dataset (Harte and Gillen~\citeyear{tcd}) consists of high-quality audio and video footage of 62 speakers reading a total of 6913 phonetically rich sentences without apparent head movement. 
The video footage is recorded from two angles: straight on and $30^{\circ}$. MODALITY dataset (Czyzewski et al.~\citeyear{czyzewski2017audio}) provides facial depth information for further analysis by using the Time-of-Flight camera. 
The employed camera model is SoftKinetic DepthSense 325, which delivers the depth data at 60 frames per second with a spatial resolution of $320 \times 240$ pixels. Besides depth recordings, the 3D data can be retrieved owing to stereo RGB cameras recordings available in the corpus. 
LRW dataset (Chung and Zisserman~\citeyear{chung2016lip}) consists of 500 different words spoken by hundreds of different speakers in the wild. 
There is a significant variation of head pose across this dataset---from some videos where a single speaker is talking directly at the camera, to panel debate where the speakers look at each other, leading to some videos with extreme head poses. 
Since LRW is collected from the real world accompanied by truth labels (words), the videos are short, i.e., lasting only one second, and there is no apparent head movement in such a short time duration.

 We select GRID dataset and LRW dataset to validate talking-head methods without any head movement in this paper by considering three Properties: 
 \begin{enumerate}
  \item The lighting condition of the videos varies in both two datasets comparing with the TCD-TIMIT dataset and MODALITY dataset.  
  \item  The face region occupies a relatively large part of the frame in both GRID and LRW. Besides, the dress, race, hairstyle, and speaking speed/style varies significantly from one subject to others in those two datasets comparing with other datasets.
  \item The GRID dataset provides accurate word-level and phone-level transcriptions for all videos, and the LRW dataset contains videos with 500 words, each spoken by different identities with diverse head poses.
\end{enumerate}

\subsection{Videos With Spontaneous Motions} 
\label{subsec:small_move}
In the real-world scenarios, people naturally emit head movements and emotional state when they speak, which contain nonverbal information that helps the audience comprehend the speech content (Glowinski et al.~\citeyear{glowinski2011toward}, Ginosar et al.~\citeyear{ginosar2019learning}). 
Meanwhile, human
perception is susceptible to subtle head movement in real videos. Thus, we summarize some video datasets in which the speakers are talking with moderate and natural head movements.

In the CREMA-D dataset (Cao et al.~\citeyear{crema}), 91 actors coming from a variety of different age groups and races utter 12 sentences. 
Different from other datasets, each sentence in CREMA-D is acted out by the actors multiple times with different emotions and intensities with natural head movement. 
Similarly, the RAVDESS dataset (Livingstone et al.~\citeyear{ryerson}) and MSP-IMPROV dataset (Busso et al.~\citeyear{busso2016msp}) involve creating stimulus with conflicting emotional content conveyed through speech and facial expression. 
Faceforensics++ dataset (Rossler et al.~\citeyear{facefor}) contains 1000 videos of news briefing from different reporters. 
The speakers in the videos are facing the camera with moderate and natural head movements. ObamaSet (Suwajanakorn et al.~\citeyear{suwajanakorn2017synthesizing}) consists of an abundance of video footage from President Barack Obama's weekly presidential addresses, spanning a period of eight years. His head pose changes when he speaks while keeping his persona consistent.
Because of those characteristics, ObamaSet is a suitable dataset to study the high-quality talking-head generation for a specific subject.

We select the CREMA-D dataset to study the videos with Spontaneous Motions (e.g., natural and moderate head movements, emotional expressions) since it contains relatively large amounts of subjects, which enables the generalizability of the model.
Meanwhile, the CREMA-D dataset contains videos speaking with a wide range of emotional expression, from ambiguous to prototypical emotion, subtle to extreme expression.
In order to learn the long-term spontaneous motion of a specific subject, we also perform some experiments on ObamaSet.
 
 \subsection{Videos with Apparent Head Movements} 
\label{subsec:large_move}
Datasets discussed above are either videos recorded in a lab-controlled environment or videos that the shots are relatively controlled with the subject in the center and facing the camera.
There are some more challenging datasets that contain videos with either apparent head motion or faces with extreme poses. 
VoxCeleb1 (Nagrani et al.~\citeyear{voxceleb1}) and VoxCeleb2 (Chung et al.~\citeyear{voxceleb2}) datasets together contain over one million utterances from over 6,000 speakers, extracted from videos uploaded to YouTube. 
The speakers span a wide range of different accents, professions, ethnicities, and ages.
Videos included in the datasets are shot in a large number of challenging visual and auditory environments, with variations in lighting, image quality, pose (including profiles), and motion blur.
These include speeches given to large audiences, interviews from quiet indoor studios, outdoor stadiums and red carpets, excerpts from professionally shot multimedia, and even homemade videos shot on hand-held devices.
Audio segments present in the datasets are degraded with background chatter, laughter, overlapping speech, and varying room acoustics.
The LRS3-TED dataset (Afouras et al.~\citeyear{lrs3}) is a large-scale dataset for audio-visual speech recognition tasks, which contains word-level alignment between the subtitle and the audio signal.
MELD dataset (Poria et al.~\citeyear{MELD2018}) contains about 13,000 utterances from 1,433 dialogues from the TV-series \textit{Friends}, which are annotated with emotion and sentiment labels.

To exam the talking-head generation methods with the videos ``in the wild,'' we select VoxCeleb2 and LRS3-TED as our benchmark datasets because of the following two properties:
 \begin{enumerate}
  \item  Since VoxCeleb2 is collected ``in the wild'', the speech segments are corrupted with real-world noise, including laughter, cross-talk, channel effects, music, and other sounds. 
  The dataset is also multilingual, with speeches from speakers of 145 different nationalities, covering a wide range of accents, ages, ethnicities, and languages, enabling the generalizability of the generative model. 
  Meanwhile, it contains over a million utterances from over 6,000 speakers, which is several times larger than other publicly available audio-visual datasets.
  \item  The LRS3-TED dataset consists of over 400 hours of videos, speaking by more than 5,000 identities, with a significant variety of pose, expressions, lighting, and camera positions, along with the corresponding subtitles and word alignment boundaries.
\end{enumerate}

\subsection{Benchmark Data Pre-possessing Protocol}
\label{subsec:preprocess}

Tab.~\ref{Table:dataset} lists many existing datasets and we select six datasets as our benchmark datasets, which can be categorize into 4 categories (Tab.~\ref{Table:selected_data}). 
However, those datasets are not calibrated (e.g., scaling, cropping, face tracker, and head position).
LRW, VoxCeleb2, and LRS3-TED are videos that is pre-processed with certain protocols by using the face detector and tracker proposed in Nagrani et al.~\citeyearpar{voxceleb1}.
This pre-processing method is not publicly available, leading to a problem that the generative models trained on these datasets may not be able to generate videos with samples outside of these datasets due to the lack of the pre-processing. 
In this section, we introduce a uniformed pre-possessing protocol so that the generative models trained on one dataset can be easily transferred to test videos outside of the dataset.
The key stages of the pipeline are:  

\noindent \textbf{Face tracking.} \quad Given a video with a length of $T$, the state-of-the-art face detector (Bulat et al.~\citeyear{bulat2017far}) is used to detect the 2D facial landmarks in every frame of the video. 
Then we calculate the center points ($x_{1:T}, y_{1:T}$) of the landmarks in eye area over the frames and apply a 1D smooth operation on the sequence $x_{1:T}$ and $y_{1:T}$, respectively. 
The smooth operation is performed by a convolution on the input $x_{1:T}$ and a hanning window $w$ (Essenwanger et al.~\citeyear{essenwanger1986elements}), which is:
\begin{equation}
\centering
\begin{aligned}
\overline{x}_i = \sum_{j=0}^{N-1}x_{i + j - \frac{N-1}{2}}\frac{w_j}{\sum_{j=0}^{N-1}w_j} \enspace ,
\end{aligned}
\label{eq:smooth}    
\end{equation}
where N is the window size (we set N =11 in this paper), and $w$ is the hanning window, where
\begin{equation}
\centering
\begin{aligned}
w_j = 0.5 -0.5 cos(\frac{2\pi j }{N-1}), 0\leq j \leq N-1      \enspace .
\end{aligned}
\label{eq:smooth2}    
\end{equation}
This step can track the face with a smooth transition between adjacent frames when the head moves.

\noindent \textbf{Face cropping.} \quad The boundaries of face region in each image are calculated based on the 2D landmarks detected in the tracking step. Then we calculate the mean length $l$ of the face region in each video, and obtain the top-left coordinate sequence by:
\begin{equation}
\centering
\begin{aligned}
(\overline{x}_{1:T}, \overline{y}_{1:T}) =   (\overline{x}_{1:T} - r_1 \times l, \overline{y}_{1:T} - r_2 \times l ) \enspace,
\end{aligned}
\label{eq:training}    
\end{equation}
where $r_1$ and $r_2$ are hyper-parameters. We set $r_1$ and $r_2$ equal to $\frac{10}{9}$ and $\frac{8}{9}$, respectively. Then we crop the square region with the side length of $\frac{41}{18}$. This step decides the cropping region dynamically to migrate the scaling problem caused by camera position, with the face region occupying a relatively large part of the frame.

\subsection{Benchmark Dataset Property Distribution}
\label{subsec:data_dis}
Most recent audio-visual datasets capitalize on the large amount of video data collected from wild resources, including news, interviews, talks, etc. For example, VoxCeleb2, LRS3, LRW, and ObamaSet collect videos from YouTube, TED, BBC news, and Obama’s weekly addresses, respectively.
It is hard to balance properties of different datasets due to the large scale and wild environment. Moreover, some properties may lead to different generation performance.
In this section, we analyze the distribution of several properties across different datasets and provide some information about the datasets that have not been addressed by previous papers.

\begin{figure*}
  \includegraphics[width=0.98 \linewidth]{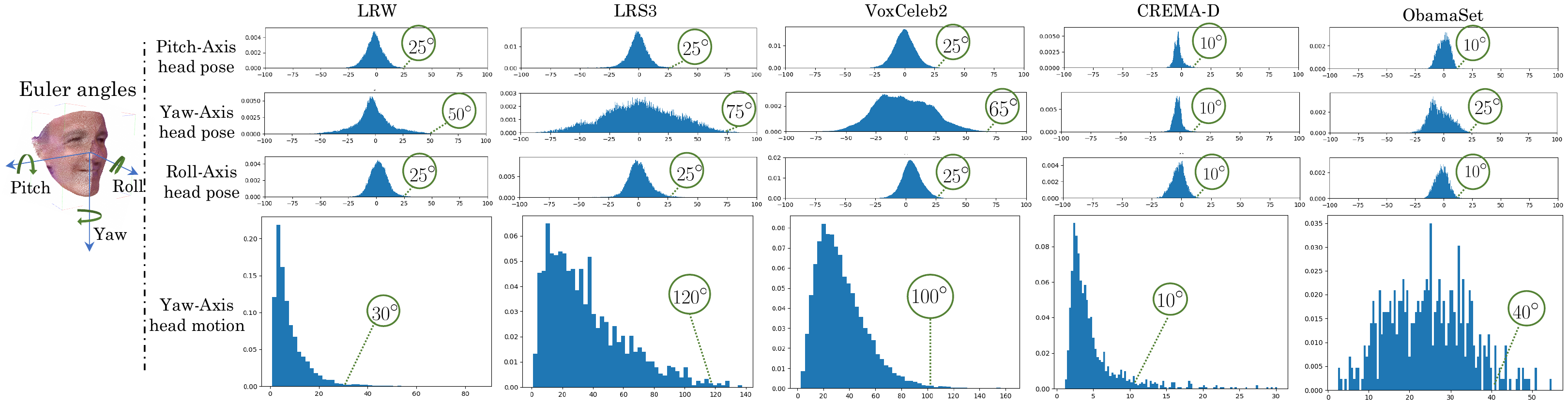}
\caption{The left column indicates the Euler angler system. On the right side, the first three rows show the distribution of head poses across different datasets in Pitch-Axis, Yaw-Axis, and Roll-Axis, respectively. The last row shows the distribution of head motion across different datasets. All the X-axis, Y-axis are the degree and ratio, respectively.}
\label{fig:pose_motion_distribution}       
\end{figure*}

\noindent \textbf{Head pose.} \quad Head pose is one of the essential attributes of talking-head videos. 
However, most previous works only consider facial animation with a fixed frontal head pose (Chung et al.~\citeyear{jamaludin2019you}; Song et al.~\citeyear{song2018talking}). 
Recent works start to model it by either relying on sparse mapping (Zakharov et al.~\citeyear{zakharov2019few}; Wang et al.~\citeyear{wang2019few}), dense mapping (Siarohin et al.~\citeyear{NIPS2019_8935}) or modeling from a short video clip (Yi et al.~\citeyear{yi2020audio}). 
Moreover, they evaluate their performance on different datasets, making it hard to compare their performance. 
To better understand the difficulty of each benchmark dataset in terms of head pose, we plot the Euler angle distribution of the head poses. 
Specifically, we extract the 3D facial landmarks from each video frame using a tool proposed in Bulat and Tzimiropoulos~\citeyearpar{bulat2017far}.
Furthermore, we register the extracted 3D landmark with a canonical 3D landmark by applying rotation and translation. We 
transfer the rotation to the Euler angle and show the overall head pose histogram in the first three rows of Fig.~\ref{fig:pose_motion_distribution}.
According to the distribution plot, it is obvious that people usually move their heads along the Yaw axis.
The LRS3-TED and VoxCeleb2 contain relatively large head motion comparing with other datasets.
Since the head movement in the roll and pitch axis is not apparent, we omit them in this paper and simplify the head motion to rotations along the yaw axis.

\noindent \textbf{Head motion.} \quad In real-world scenario, people often emit natural head movement when they are talking. 
As mentioned in Yi et al.~\citeyearpar{yi2020audio}, synthesizing talking-head videos with head motion is much more difficult than synthesizing videos with fixed head poses. To approximate the degree of head motion in a video sequence ($\mathbf{x}\equiv \mathbf{x}_{1}, ..., \mathbf{x}_{k}$),
We calculate the maximum difference of head poses in the video:
$M_{\mathbf{x}} \approx |\max(P(\mathbf{x}_{1}), ..., P(\mathbf{x}_{k})) - \min(P(\mathbf{x}_{1}), ..., P(\mathbf{x}_{k}))|$, 
where the $P(\cdot)$ denotes the head pose estimation operation. This head motion estimation can only estimate head motion in a short video clip (e.g., less than 20 seconds). We plot the histograms head motion distribution across different datasets in the last row of Fig.~\ref{fig:pose_motion_distribution}, where we can find that LRS3-TED and VoxCeleb2 contain much more diverse head motions.


\section{Evaluation Metrics}
\label{sec:evaluation}

Evaluating the visual quality and naturalness of synthesized videos, in particular regarding the human face, is challenging.
Recall that in Sect.~\ref{sec:generation}, we discussed 
four criteria for evaluating talking-head generation algorithms: identity preserving, visual quality, lip synchronization, and natural, spontaneous motion.
In this section, we will discuss 
related evaluations and benchmarks concerning the four criteria.
Firstly, we provide several evaluation metrics, accessing the four desired properties of talking-head videos followed by an evaluation of whether a given measure or a family of measures are able to access them.
Then, if the existing measures can not judge these properties, we will introduce new evaluation metrics.

\subsection{Identity Preserving} 
\label{subsec:eval_identity}

\begin{figure*}
  \includegraphics[width=0.98 \linewidth]{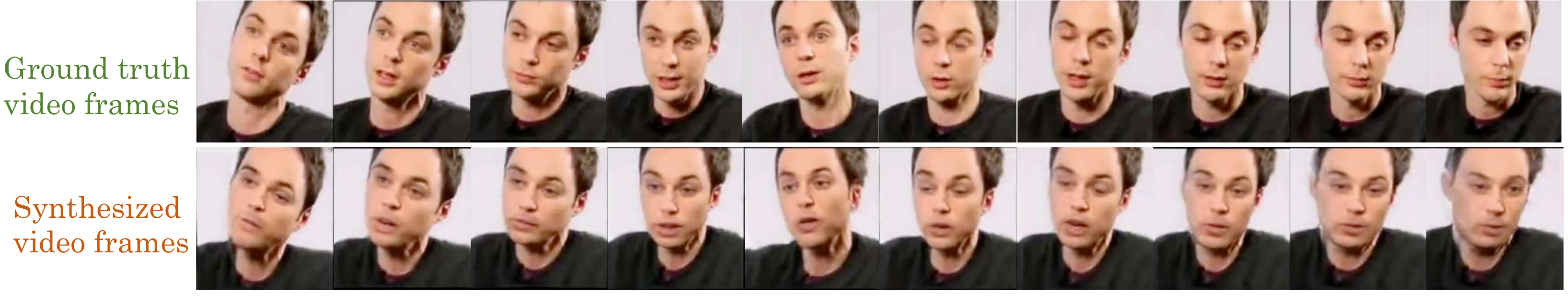}
\caption{The video frames with changing facial appearance. The second row shows the results synthesized by our baseline on VoxCeleb2 testing set.}
\label{fig:loss_identity}      
\end{figure*}

In real-world scenarios, human perception is sensitive to subtle appearance changes in real videos. 
Fig.~\ref{fig:loss_identity} shows one example result synthesized by our baseline model, where the identity keeps changing over time as the head pose changes. 
To evaluate the identity-preserving performance, Jamaludin et al.~\citeyearpar{jamaludin2019you} use the embedding distance of the generated video frames and the ground truth using a pre-trained VGG Face classification network (VGGFaceNet, Parkhi et al.~\citeyear{parkhi2015deep}) to measure the identity distance since it is trained with a triplet loss. 
 ArcFace classification network (Deng et al.~\citeyear{deng2019arcface}) has been adopted in Zakharov et al.~\citeyearpar{zakharov2019few}. 
 Specifically, they obtain the embedding vectors in latent space using the state-of-the-art face recognition network (ArcFace, Deng et al.~\citeyear{deng2019arcface}).
ArcFace is trained with arc loss, which consists of two parts---Softmax loss and additive angular margin.
 Then they compute the cosine distance of two vectors for measuring identity mismatch.
 
  \begin{figure}[t]
  \includegraphics[width=0.98 \linewidth]{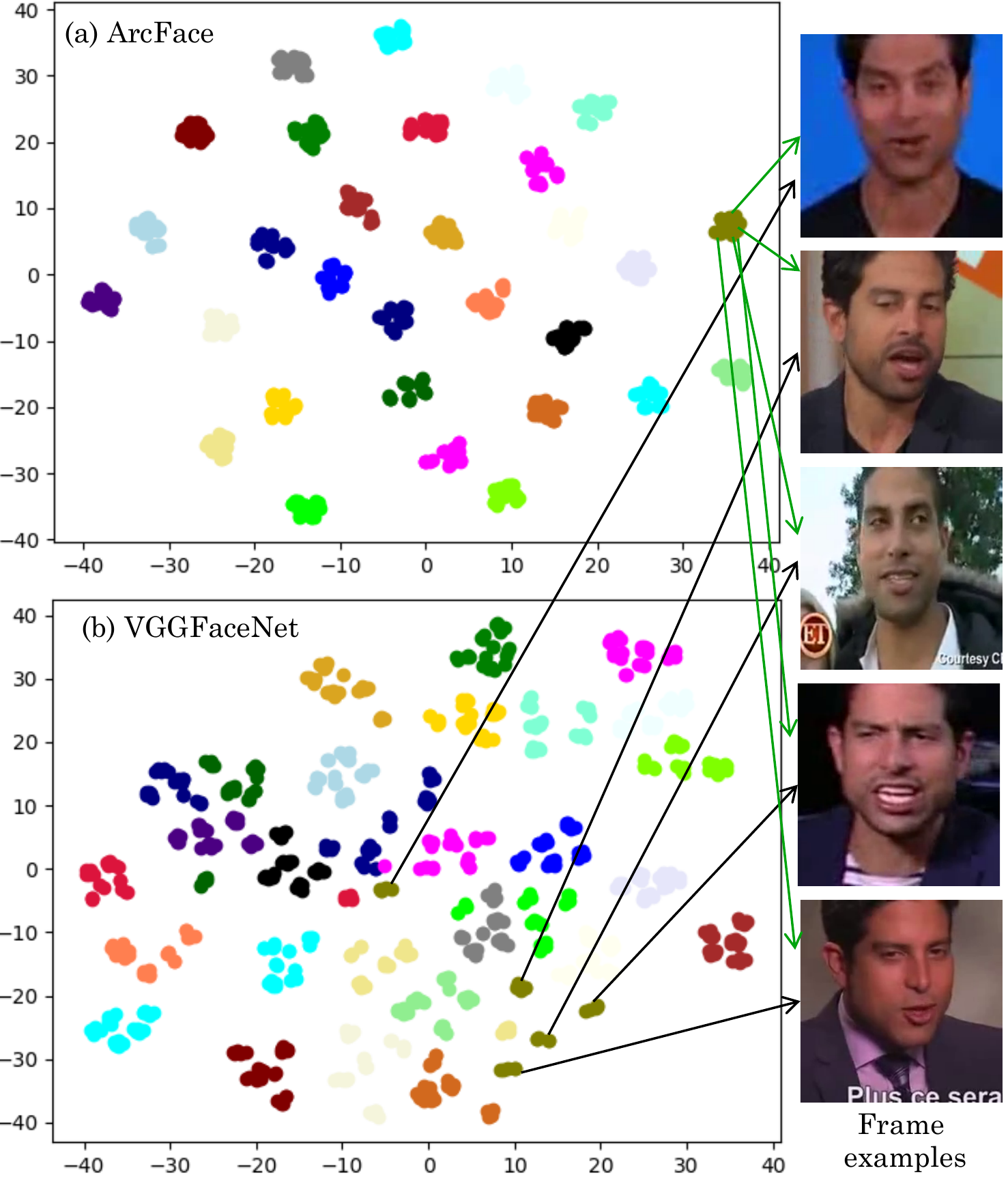}
\caption{The t-SNE plot of identity features of random frames from VoxCeleb2 testing set. The features are extracted by VGGFaceNet and ArcFace, respectively. Frames corresponding to the same subject have the same color.}
\label{fig:tsne_embedding}       
\end{figure}

 To compare those two different embedding methods, we use t-SNE (Maaten and Hinton~\citeyear{maaten2008visualizing}) to visualize the extracted feature vectors of video frames sampled from VoxCeleb2 (see Fig.~\ref{fig:tsne_embedding}). 
 Specifically, we randomly sample 150 videos from 30 identities (each with five different videos). 
 While the identity is the same in each 5-videos set, the hairstyle, lightning, age, background, video quality, and head pose varies among those videos since those videos are recorded at different times under different conditions. 
 In each video, we randomly sample 5 frames, and there are 750 images in total. From the t-SNE plot, we find that ArcFace (Fig.~\ref{fig:tsne_embedding}a) is more robust to noise (e.g., hairstyle, lighting, and video quality) comparing with VGGFaceNet (Fig.~\ref{fig:tsne_embedding}b). 
 We attribute this to the Additive Angular Margin Loss (ArcFace) proposed in Deng et al.~\citeyearpar{deng2019arcface} since it simultaneously enhances the intra-class compactness and inter-class discrepancy.
 Based on the observation that ArcFace has better inter-class discrepancy ability, we use ArcSim---the cosine distance between the two image features extracted by ArcFace to measure the identity similarity between two images.
 
\subsection{Visual Quality}
\label{subsec:visual_metrics}
In talking-head video frames, while there are typical deformations caused by facial expressions (e.g., talking, emotional expression, natural head movements), other distortions present. For example, the artifacts due to the method itself (e.g., is agnostic to mode collapse).
Thus, an ideal realism measure should be invariant to these facial deformations but sensitive to the artifacts.
For instance, the score of a synthesized talking-head image frame should not change much if its generated faces are with different facial expressions or head movements. 

Reconstruction loss can stabilize the adversarial training, and it is widely used for talking-head generation.
It is, therefore, natural to evaluate synthesized images using a reconstruction error measure (e.g., Mean squared error) computed on the testing set.
However, reconstruction error metrics are sensitive to misalignment between the synthesized frames and the ground truth, which is hard to avoid in talking-head video generation tasks due to some noisy movements in the wild videos. If there exists a corresponding ground truth frame, we can use the Peak Signal-to-Noise Ratio (PSNR) to access the synthesized image quality. 
However, these metrics ignore the features of human image perception. The SSIM (Wang et al.~\citeyear{wang2004image}) is a well-characterized metric that measures perceptual similarity, aiming to discount aspects of an image that are not important for human perception. 
It compares corresponding pixels and their neighborhoods in the synthesized image and ground truth, using three properties---contrast, luminance, and structure. 

These measures, however, may favor methods that generate well-looking images albeit with low diversity.
Inception score (IS), proposed by Salimans et al.~\citeyearpar{salimans2016improved}, shows a reasonable correlation with the visual quality and diversity of synthesized images. 
And more recently, FID is introduced by Heusel et al.~\citeyearpar{heusel2017gans}, which performs better than IS in terms of robustness, discriminability, and computational efficiency. 
FID uses a specific layer of InceptionNet (Szegedy et al.~\citeyear{szegedy2016rethinking}) to compute the latent feature of the input image. 
If we regard the embedding network as a continuous multivariate Gaussian, we can obtain the FID score by computing the Wasserstein-2 distance between these two Gaussian distributions. It has been shown that FID can measure the distance between synthetic and real data distributions, and is consistent with human perception evaluation. 

To measure the loss of sharpness during generation, we use a non-reference measure---cumulative probability blur detection (CPBD) (Narvekar and Karam~\citeyear{narvekar2009no}). 
By integrating the concept of cumulative probability blur detection with the just noticeable blur into a probability summation model, CPBD assesses image sharpness from a perceptual perspective.

In this paper, we choose SSIM and FID to measure the image-level visual quality of the synthesized video frames and use CPBD to assess the sharpness.


\subsection{Semantic-level Lip Synchronization}
\label{subsec:LRSD}
  \begin{figure}[t]
  \includegraphics[width=0.98 \linewidth]{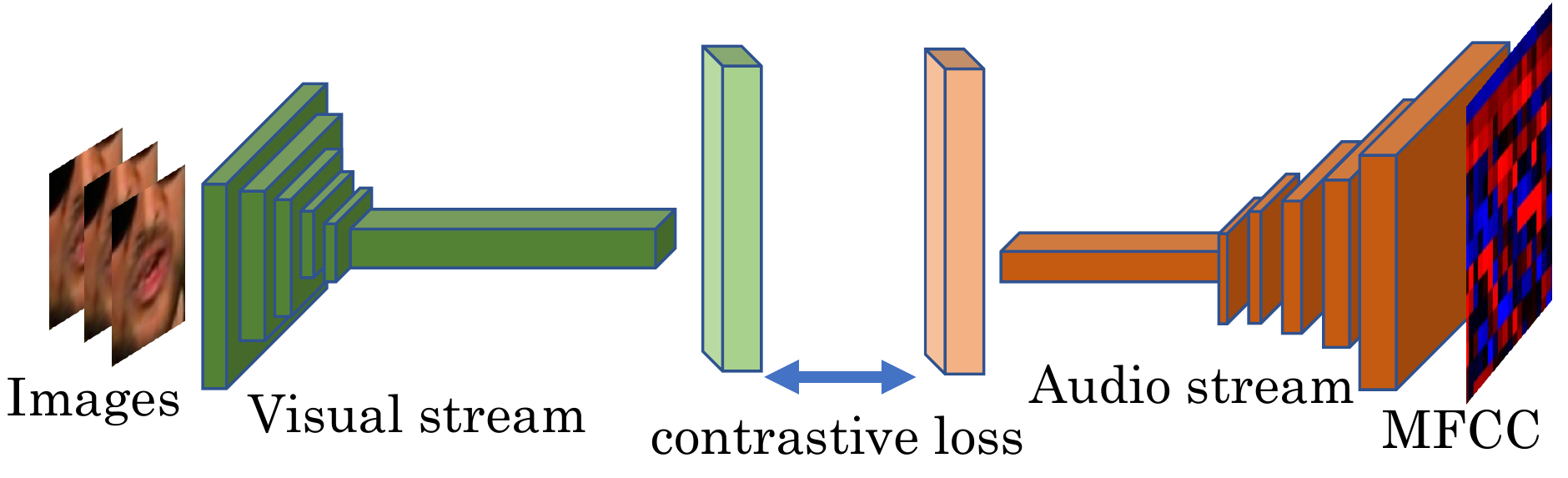}
\caption{SyncNet architecture proposed by Chung and Zisserman~\citeyearpar{chung2016lip}. Both streams are trained simultaneously.}
\label{fig:syncnet}       
\end{figure}

While there exist some prominent metrics can be used to determine frame/video quality, they fail to reflect other essential aspects of the video, such as audio-visual synchronization.
Landmark Distance metric (LMD) (Chen et al.~\citeyear{chen2018lip}) computes the Euclidean distance of the lip region landmarks between the synthesized video frames and ground truth frames.
The LMD estimates the lip shape accuracy, which can represent the lip-sync between synthesized video frames and audio-signal. 
However, LMD does not evaluate the lip-synchronization directly, and the evaluation system is not related to how humans perceive lip synchronization.
Meanwhile, LMD can not reflect the details of lip movements since there are only 20 sparse points around the lip region when we calculate the LMD. 
Zhou et al.~\citeyearpar{zhou2019talking} utilize the audio-visual retrieval protocol to evaluate the closeness between the audio and visual features. However, the retrieving system may bias to appearance (e.g., identities or head poses). 
They also perform a word-level audio-visual speech recognition evaluation on the embedded audio-visual representations. Their results show that their audio-visual embedding features outperform the features extracted by Chung and Zisserman~\citeyearpar{chung2016lip}. 
However, this step is evaluating the embedding ability of its network instead of evaluating the synthesized video quality. 
Chung et al.~\citeyearpar{chung2016out} propose a two-stream SyncNet (Fig.~\ref{fig:syncnet}) to encode the audio feature and visual feature, then use the contrastive loss to train the matching problem. 
The false pairs in the SyncNet training are just obtained by randomly shifting, and all the output scores are related to offset.
While the SyncNet can output correct synchronization error (offset) between the input audio and visual signal, it may not perform well in videos that the lip regions are intermittently synced with the audio (e.g., some frames are synced with audio while others are not), which is a general problem of synthesized video.

When humans look at a talking-head video, we would unintentionally use the semantic information to judge if the audio is synced with the visual. 
For example, it is easier for us to tell if the audio is synced with the visual when we know the language. 
Thus, in this paper, we propose a lip-synchronization evaluation metric---Lip-Reading Similarity Distance (LRSD), like human perceptual judgments by incorporating the semantic-level lipreading. 
Given a synthesized video clip $\hat{x}$ and paired ground truth video clip $x$, the LRSD is obtained by:
\begin{equation}
\centering
\begin{aligned}
   \text{LRSD}(x,\hat{x}) = ||\phi(x), \phi(\hat{x})||_2^2 \enspace,
\end{aligned}
\label{eq:LRSD}    
\end{equation}
where the $\phi$ is a spatial-temporal lipreading network. 
Although there are many lipreading networks proposed in recent years, they are either limited to frontal face videos (Assael et al.~\citeyear{assael2016lipnet}), or publicly unavailable (Zhang et al.~\citeyear{zhang2019spatio}; Chung and Zisserman~\citeyear{chung2016lip}; Chung and Zisserman~\citeyear{Chung17a}). 
Meanwhile, most of the existing lipreading methods do not perform well on videos outside the dataset, not to mention assessing the similarity between synthesized videos and real videos. 
Thus, we propose an easy but effective multi-view lipreading network, which is trained on LRS3-TED dataset and works for any video outside the LRS3-TED dataset. 
Rather than adopting RNN-seq2seq (Sutskever et al.~\citeyear{sutskever2014sequence}) structure or RNN-CTC (Graves~\citeyear{graves2006connectionist}) to handle the lipreading, we propose a spatial-temporal aware multi-view lipreading network, which outputs word-level label for the input video clip. 
The reason why we choose the word-level classification is that it learns visual features without any extra knowledge (e.g., contextual information) comparing with RNN-seq2seq or RNN-CTC models, which is consistent with how human tell if the audio is synced with visual. 
For example, if we use a strong NLP model like RNN-seq2seq, it is hard to determine that the lipreading ability is gained from visual modeling or language modeling. 

  \begin{figure*}[t]
  \includegraphics[width=0.98 \linewidth]{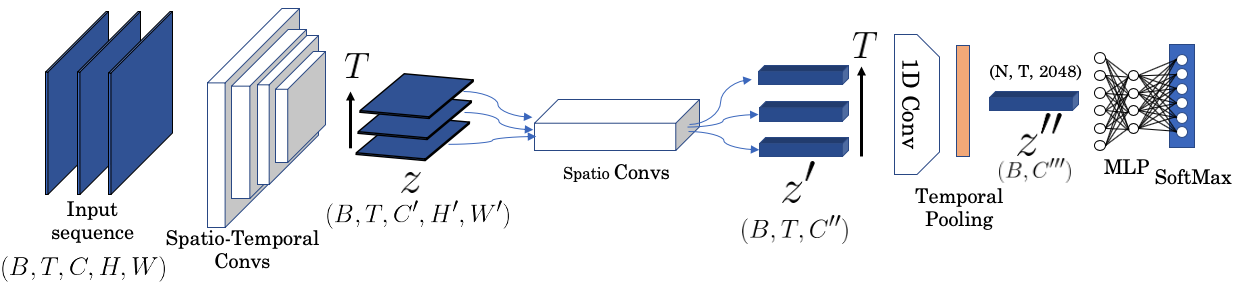}
\caption{Spatial-temporal aware network structure. Specifically, we first exploit several layers of Spatio-Temporal Convs (e.g. 3D convolutional layer) to extract feature from input sequence, and then apply Spatio Convs (e.g. 2D Convolutional layer) for further encoding. A multi-layer perceptron (MLP) is added after Spatio Convs for classification. }
\label{fig:lipnet1}       
\end{figure*}

Specifically, our lipreading network (Fig.~\ref{fig:lipnet1}) consists of three modules: a spatial-temporal aware feature extraction, a spatial feature refine module, and a fusion module. 
To capture the spatial-temporal characteristics of the lip dynamics, we adopt $3D$ convolutions on the input sequence. 
As for the following spatial feature refine module, ResNet-18 is adopted in consideration of the computational costs. 
After the refine module, we obtain the visual feature map $z \in \mathbb{R}^{B\times T \times C' }$, where $B, T$, and $C'$ denote batch size, input sequence length, and the number of channels, respectively. 
In the spatial-temporal fusion module, we apply a $1D$ convolution and a global average pooling layer to aggregate the spatial feature from $z$, and obtain a visual feature $z' \in \mathbb{R}^{B\times C''}$, where $C{''}$ is the number of output channels. 
Then the $z'$ is passed through a multi-layer perception network with Softmax to predict the word-level label. To ease the training of lipreading network, our lexicon is obtained by
selecting the 300 most frequently occurring words in the LRS3-TED training set. It is worth mentioning that we also tried to use metric-learning to refine the lipreading network with ArcLoss and Triplet loss (Chechik et al.~\citeyear{chechik2010large}). 
However, it takes a very long time to converge the metric-training step, and we remove the metric-learning step.

  \begin{figure}[t]
  \includegraphics[width=0.98 \linewidth]{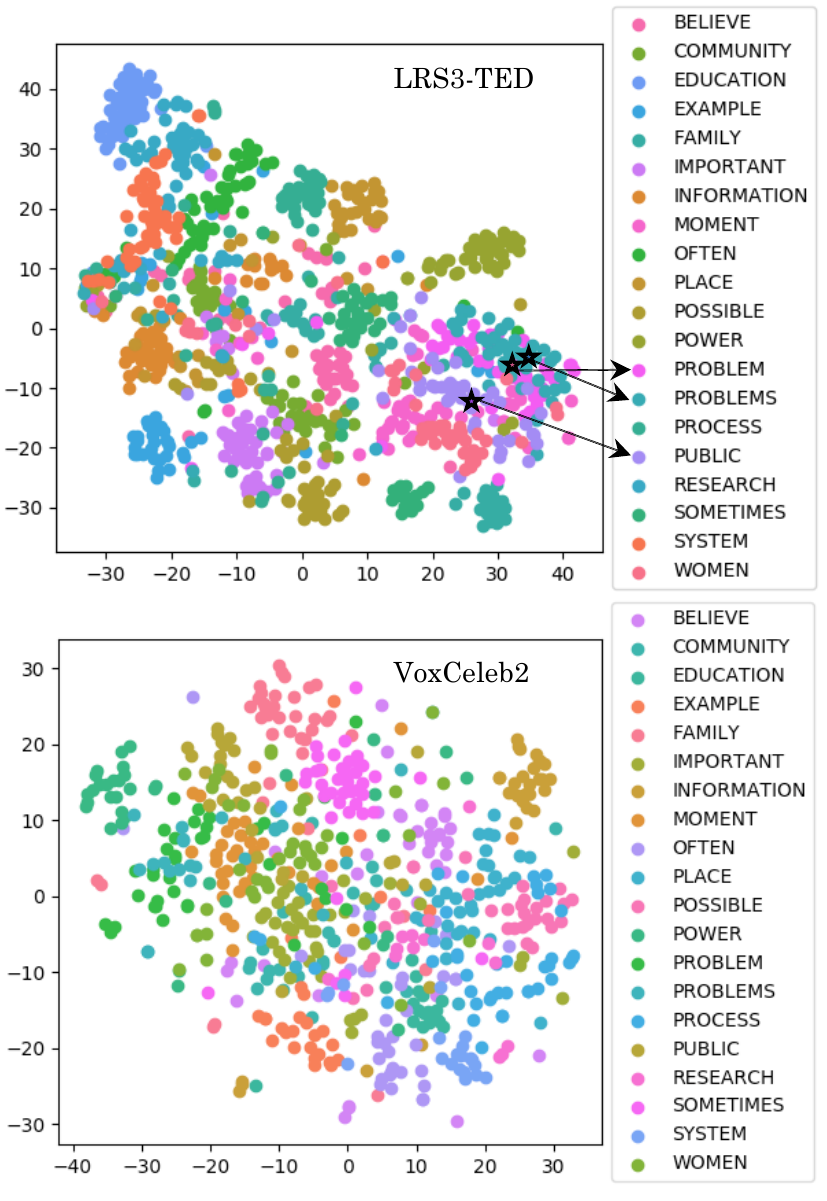}
\caption{The t-SNE plots of semantic-level visual features of random videos from the LRS3-TED testing set and VoxCeleb2 testing set. Videos corresponding to the same word have the same color.}
\label{fig:LRSD_gt}       
\end{figure}

To demonstrate the visual feature extraction ability of our lipreading network, we show the lipreading results on the testing set of LRS3-TED and VoxCeleb2 datasets in Fig.~\ref{fig:LRSD_gt}.
In order to show the inter-class discrepancy ability of the lipreading feature, we randomly select 20 words from our vocabulary, each with 30 video clips in each testing set, and visualize their visual features.
We can find that the features of words with similar visemes are closer than other features.
For example, the visual representations of `PROBLEM' and `PROBLEMS' are almost overlapped, and they are close to `PUBLIC' since the visemes of 'PUBLIC' and 'PROBLEM' are similar.
We also show the lipreading accuracy in Tab.~\ref{tab:lipread}, from where we can find that our lipreading network achieves $42.46\%$ top-1 classification accuracy on the VoxCeleb2 testing set (note that our classifier is trained on LRS3-TED dataset). We also report the top-5, top-10, and top-20 classification accuracy since we only need to compute the semantic-level lip-sync similarity for the LRSD score.
From the t-SNE plot and classification accuracy, we can find that our lipreading network can extract the semantic-level spatial-temporal features from the input video sequence, and there are clear margins between the extracted features when they do not belong to a same word.

\begin{table*}
    \centering
    \scriptsize
    \caption{The results across different methods and different datasets using four existing evaluation metrics. For ArcSim, SSIM, and CPBD, the higher the better. For FID, the lower the better.}
    \setlength\tabcolsep{0.8pt}
    \begin{tabular}{l|c|c|c|c|c|c|c|c|c|c|c|c|c|c|c|c}
        \toprule
        \hline
        \multirow{2}{*}{Methods} & \multicolumn{4}{|c|}{VoxCeleb2} & \multicolumn{4}{|c|}{LRS3-TED} & \multicolumn{4}{|c|}{LRW} & 
        \multicolumn{4}{|c}{GRID}\\\cline{2-17}
        & ArcSim & SSIM & CPBD & FID & ArcSim & SSIM & CPBD & FID & ArcSim & SSIM & CPBD & FID & ArcSim & SSIM & CPBD & FID \\\hline
        baseline & 0.40 & \textbf{0.69} & \textbf{0.22} & \textbf{72.79} & 0.36 & 0.61 & 0.18 & 133.75 & 0.63 & \textbf{0.67} & \textbf{0.34} & \textbf{87.35} & 0.26 & 0.81 & 0.33 & 68.04 \\
        Wang et al.~\citeyearpar{wang2019few} & 0.41 & 0.67 & 0.18 & 104.21 & 0.43 & \textbf{0.62} & 0.18 & \textbf{120.39} & 0.64 & 0.35 & 0.28 & 239.40 & \textbf{0.92} & \textbf{0.83} & \textbf{0.33} & \textbf{37.98} \\
        Wiles et al.~\citeyearpar{x2face} & 0.36 & 0.60 & 0.21 & 265.10 & 0.36 & 0.61 & \textbf{0.23} & 162.40 & 0.60 & 0.63 & 0.22 & 206.16 & 0.82 & 0.81 & 0.29 & 70.83 \\
        Zakharov et al.~\citeyearpar{zakharov2019few} & 0.30 & 0.32 & 0.10 & 141.36 & 0.31 & 0.47 & 0.10 & 153.34 &0.47 & 0.42 & 0.11 & 105.97 & 0.70 & 0.54 & 0.19 & 100.14 \\
        Jamaludin et al.~\citeyearpar{jamaludin2019you} & 0.55 & 0.21 & 0.18 & 296.27 & 0.58 & 0.39 & 0.14 & 279.74 & 0.72 & 0.34 & 0.21 & 197.05 & 0.77 & 0.41 & 0.22 & 264.78 \\
        Chen et al.~\citeyearpar{chen2019hierarchical} & \textbf{0.59} & 0.21 & 0.04 & 330.88 & \textbf{0.62} & 0.39 & 0.03 & 266.69 & \textbf{0.79} & 0.38 & 0.07 & 189.59 & 0.85 & 0.41 & 0.08 & 218.37 \\
    \end{tabular}
    \label{tab:overall}
\end{table*} 
\begin{figure}
  \includegraphics[width=0.98 \linewidth]{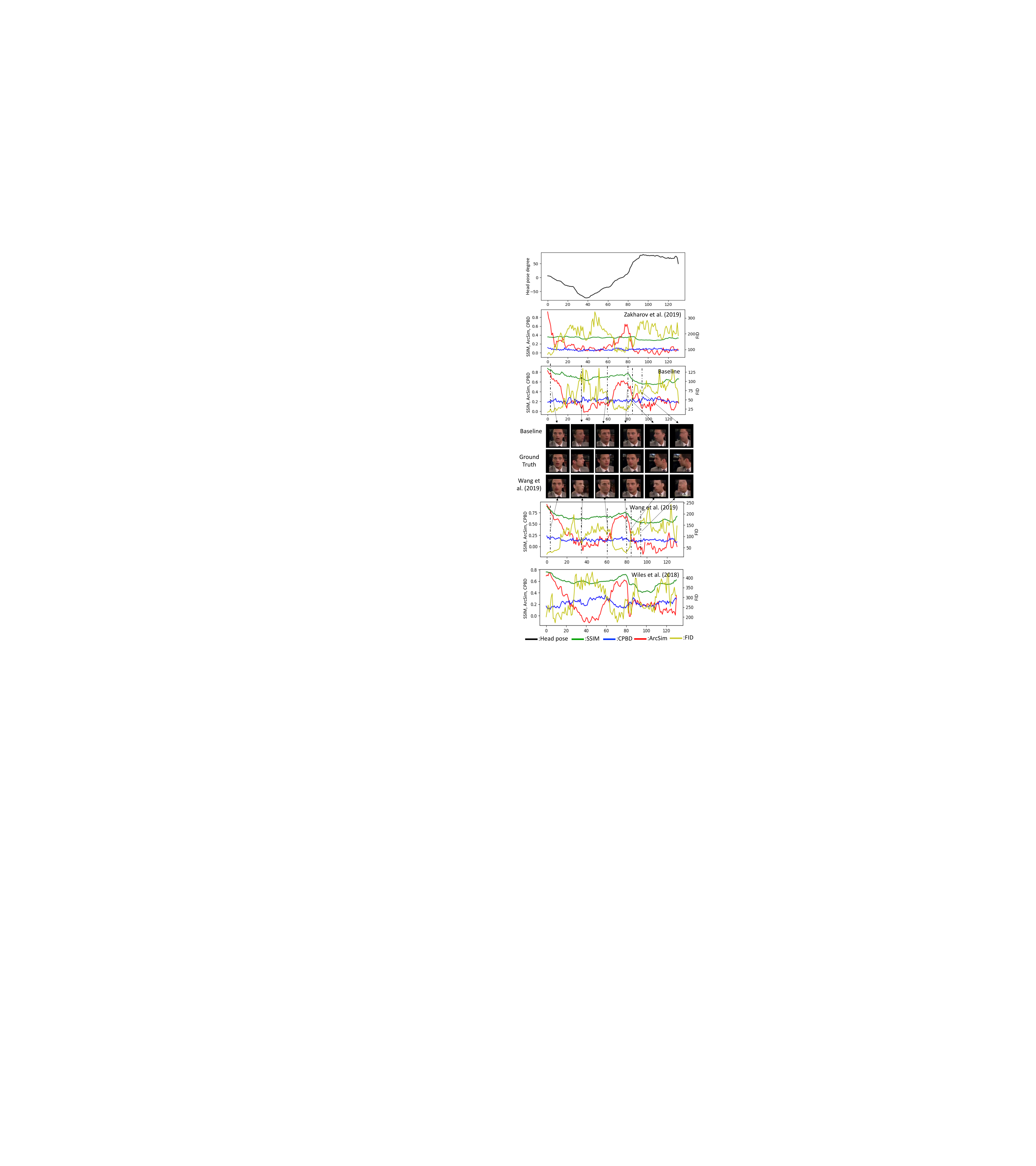}
\caption{Evolution of visual quality and metric scores on generated images versus the head pose in a single video. The X-axis is the frame id.}
\label{fig:trend_pose}       
\end{figure}

\begin{figure}[t]
  \includegraphics[width=0.9 \linewidth]{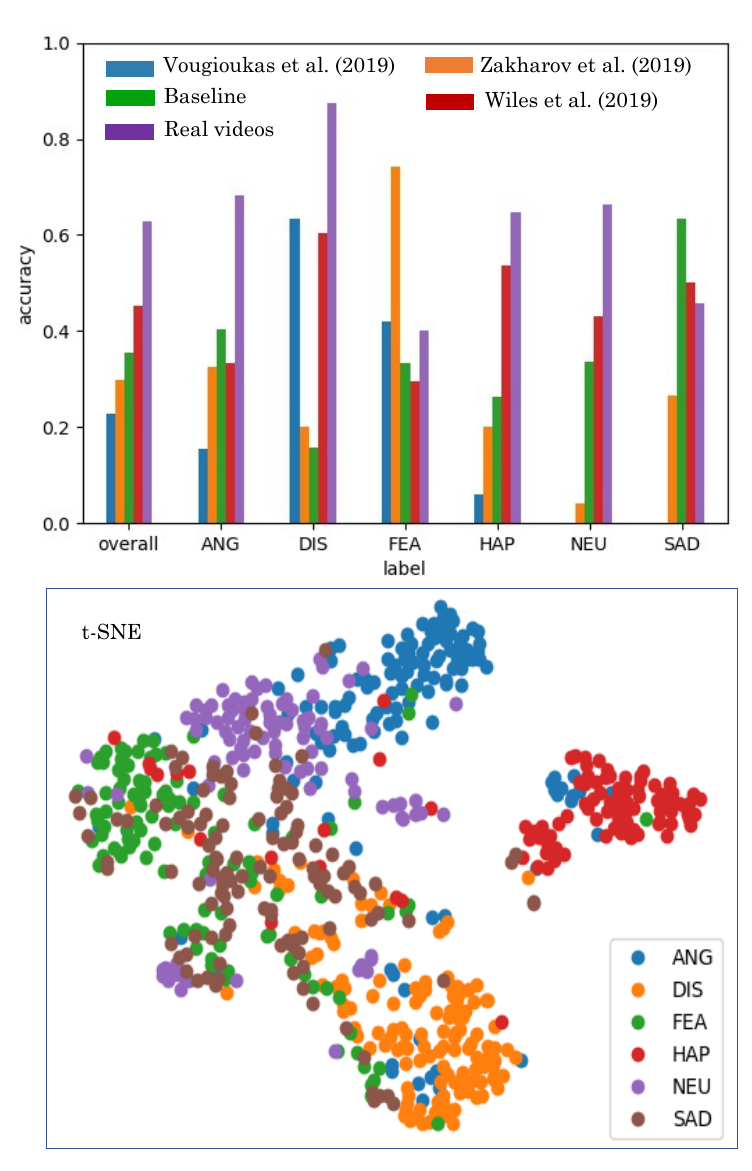}
\caption{The upper row shows the video emotion classifier's performance on the CREMA-D testing set. The X-axis and Y-axis are emotion labels and classification accuracy, respectively. The second row is the t-SNE plot of the emotion encoding of random video samples from the CREMA-D testing set. Videos corresponding to the same emotion label have the same color.}
\label{fig:emotion_cls}       
\end{figure}

\subsection{Spontaneous Motion}
\label{subsec:sec4_motion}
Investigating the generation of spontaneous expressions is also important since it is one of the main factors that affect our perception of how natural a video looks. 
Vougioukas et al.~\citeyearpar{vougioukas2019realistic} calculate the eye aspect ratio (EAR) to detect the occurrence of blinks in a video and evaluate the blink distribution between real videos and synthesized videos in different datasets. However, this eye blink distribution can not reflect if the motion is natural and smooth. Recently, the optical flow has become a common tool to visualize the facial motion between consecutive frames (Song et al.~\citeyear{song2018talking}; Vougioukas et al.~\citeyear{vougioukas2019realistic}).
However, evaluating the motion quality by optical flow with human vision is expensive and cumbersome, biased (e.g., depends on the quality of the subjective), difficult to reproduce, thus, does not fully reflect the capacity of the models. 
\begin{figure*}[t]
  \includegraphics[width=0.98 \linewidth]{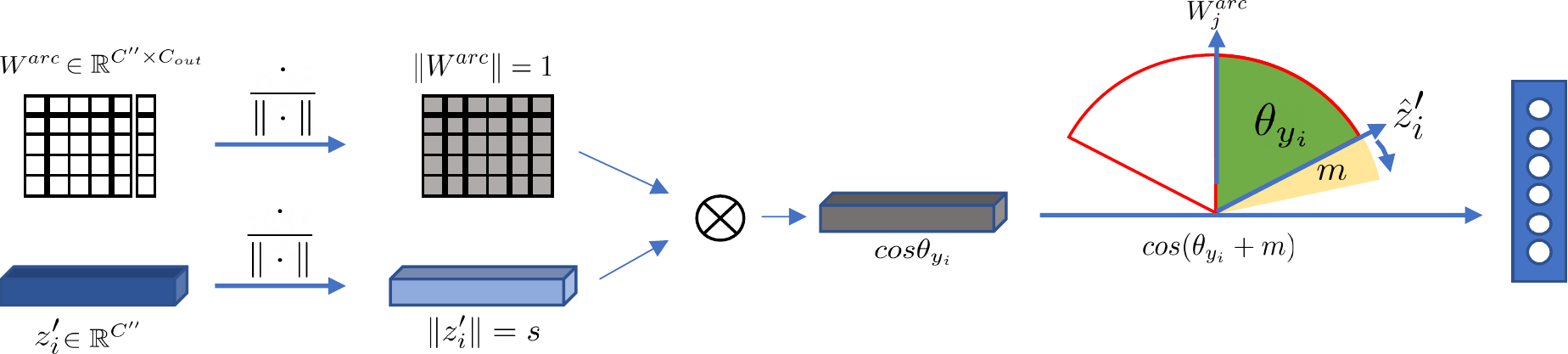}
\caption{The ArcLoss module for our metric learning step. Both matrix parameters $W^{arc}$ and encoded video vector $z_i'$ are pre-processed by $l_2$ normalization in the first step. Then the normalized results are multiplied and transferred to hypersphere distribution space by timing $\cos{(\theta_{y_j}+m)}$,where $m$ refers to margin that enhance intra-class compactness and inter-class discrepancy.}
\label{fig:arcloss}       
\end{figure*}
We study the spontaneous motion problem by considering the emotional expression during talking-head video generation.
To evaluate the quality of synthesized spontaneous motion (emotional expression), we introduce a new emotion similarity distance (ESD).
We first train a spatial-temporal convolution network (Fig.~\ref{fig:lipnet1}) to classify emotions of video clips in the CREMA-D training set.
The upper row of Fig.~\ref{fig:emotion_cls} shows the video emotion classification accuracy on the CREMA-D testing set.
According to the user studies in Cao et al.~\citeyearpar{crema}, the human recognition of intended emotion on the CREDMA-D dataset are $58.2\%$ (visual-only) and $63.6\%$ (audio-visual), respectively.
Our video-level emotion classifier achieves $62.9\%$ testing accuracy on real videos without audio, which is better than individual human raters (visual-only).
Interestingly, our classifier achieved lower scores in videos with sad and fear, which is consistent with the visual-only human performance reported in Cao et al.~\citeyearpar{crema}.

However, though the classification network introduced in Fig.~\ref{fig:lipnet1} demonstrates its capability to classify different emotions, the model only applies a Softmax layer following the final linear transform. As illustrated in Deng et al.~\citeyearpar{deng2019arcface}, the learned features from Softmax layer are not discriminative enough for the open-set recognition problem, which is high likely to influence the generalization of our model to videos generated by various algorithms. Moreover, the model with the Softmax layer does not explicitly optimize the feature embedding with higher intra-class similarity and inter-class diversity. Therefore, we follow the configuration in Deng et al.~\citeyearpar{deng2019arcface} in which we replace the last Softmax layer of our model with ArcLoss to refine the feature embedding. Fig.~\ref{fig:arcloss} shows the details of ArcLoss after the model encodes one video to $z'_i$ with $C''$ channels. $W^{arc}$ represents matrix parameters to transform $z'_i$ linearly. Instead of the simple multiplication in Softmax, the module first normalizes both $z'_i$ and $W_{arc}$ with their length equal to $s$ and $1$, respectively, by $l_2$ normalisation, and then transform them as $z'_iW^{arc}=\|z'_i\|\|W^{arc}_j\|cos\theta_j$. The generated embedding features are thus distributed on a hypersphere. An margin penalty $m$ is added to the feature according to the ground truth label, which further enforces $W^{arc}_j$ and $z'_i$ to enhance the intra-class compactness and inter-class discrepancy. Then the Softmax operation is applied on the enhanced feature as:

\begin{equation}
\centering
\begin{aligned}
   L_{arc}= & -\frac{1}{N} \sum_{i=1}^N \log( \\ & \frac{e^{s(\cos(\theta_{y_i}+m))}}{e^{s(\cos(\theta_{y_i}+m))}+\sum_{j=1, j\neq y_i}^ne^{s(\cos\theta_j)}})\enspace ,
\end{aligned}
\label{eq:arcloss}    
\end{equation}
where $N$ is the total number of video samples in one batch.

After training with the metric learning step, the emotion features from the model will have a more clear margin than the feature obtained by classification network, and we extract the emotion features from real videos in CREMA-D testing set and show the t-SNE plot in Fig.~\ref{fig:emotion_cls} second row.
We can find that the video emotion features extracted by our classifier have the inter-class discrepancy ability so that features with the same label can be grouped with clear boundaries. Besides plotting the distribution of generated embedding features, we also measure the similarity between true videos and generated results qualitatively. Since the video features from our model are optimized on a hypersphere with cosine angles, it is naturally to apply cosine similarity as Emotion Similarity Distance (ESD). Therefore, after training with ArcLoss, we utilize the embedding features before the ArcLoss module to represent each input video and calculate their similarity distance as: 
\begin{equation}
\centering
\begin{aligned}
   \text{ESD}(v_i, v_j) &  = \frac{v_i\cdot v_j}{\|v_i\|\|v_j\|} \\ &
   =\frac{\sum_{k=1}^nv_i^kv_j^k}{\sqrt{\sum_{k=1}^n(v_i^k)^2}\sqrt{\sum_{k=1}^n(v_j^k)^2}}\enspace ,
\end{aligned}
\label{eq:esd}    
\end{equation}
where $i$ and $j$ are indexes for two videos respectively. The ESD result is shown in Tab.~\ref{Table:esd} and we will discuss it in the following part. 

  \begin{figure}[t]
  \includegraphics[width=0.98 \linewidth]{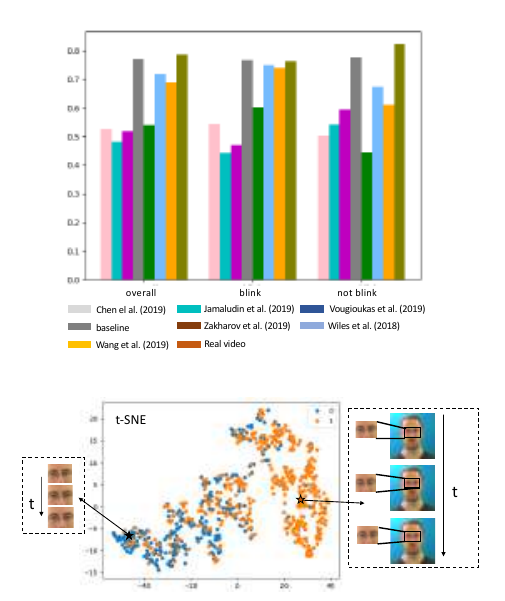}
\caption{The evaluation of blink motion. The histogram shows the performance of blink model on GRID's testing set. We evaluate the model on both original real videos in GRID and also synthesized videos by seven example methods. We also show the t-SNE plot for blink features extracted by our blink model. In the bottom figure, points with label 1,0 are belonged to blink motion, and non-blink motion, respectively.}
\label{fig:blink_accuracy}       
\end{figure}

Additionally, we also consider blink as a spontaneous motion and similar to emotion classification we evaluate quality of synthesized blink motion with the same backbone of spatial-temporal convolutional network (Fig.~\ref{fig:lipnet1}) but with separate parameters. For the reason that frames of blink motion are not annotated in talking head generation datasets, we need to create a new blink-dataset to train the blink model. We first calculate open rate of eyes in GRID dataset based on landmarks for each frame and annotate frames to be either opened eyes or closed eyes based on open-rate distribution. Then we sample slices with $t$ frames that contain the change between opened eyes and closed eyes as blink motion, and slices that only contain opened eyes or closed eyes as non-blink motion (e.g., Fig.~\ref{fig:blink_accuracy}). The blink model is trained on a subset of blink-dataset that contains 133,824 slices to recognize whether blink motion happened for each slice and evaluated on an excluded testing set with 59,253 slices of blink-dataset. As shown in Fig.~\ref{fig:blink_accuracy}, for both training and testing processes, we crop the eyes regions from each frames and input them into the network. The classification accuracy of original video in GRID's testing set is shown in the histogram in Fig.~\ref{fig:blink_accuracy} as 'real video' bar, 78.71\% for overall test set. Specifically, for blink slices in test set our model achieves 76.22\% accuracy while 82.44\% for non-blink slices. 

We also train the network with ArcLoss and extract blink features for each slices the same as what we do for emotion features. The t-SNE plot over blink features of sampled slices from test set is shown in Fig.~\ref{fig:blink_accuracy}. Although confusion slices exist for blink model, the blink features represent obvious inter-class discrepancy ability, that is non-blink motion cluster on the left and blink motion cluster on the right. Based on this observation, we introduce Blinking Similarity Distance (BSD) to better evaluate blink generation quality of synthesized videos. Similar to ESD, we calculate the cosine similarity between blink feature of ground truth videos and that of synthesized videos (same equation as Eq.~\ref{eq:esd}). A high score of BSD indicates similarity blink motion between two videos, which means both of them perform either similar blink motion or similar non-blink motion. The results of BSD will be discussed in Sect.~\ref{subsec:spontMotionExp}.

\begin{figure*}[t]
 \includegraphics[width=0.98 \linewidth]{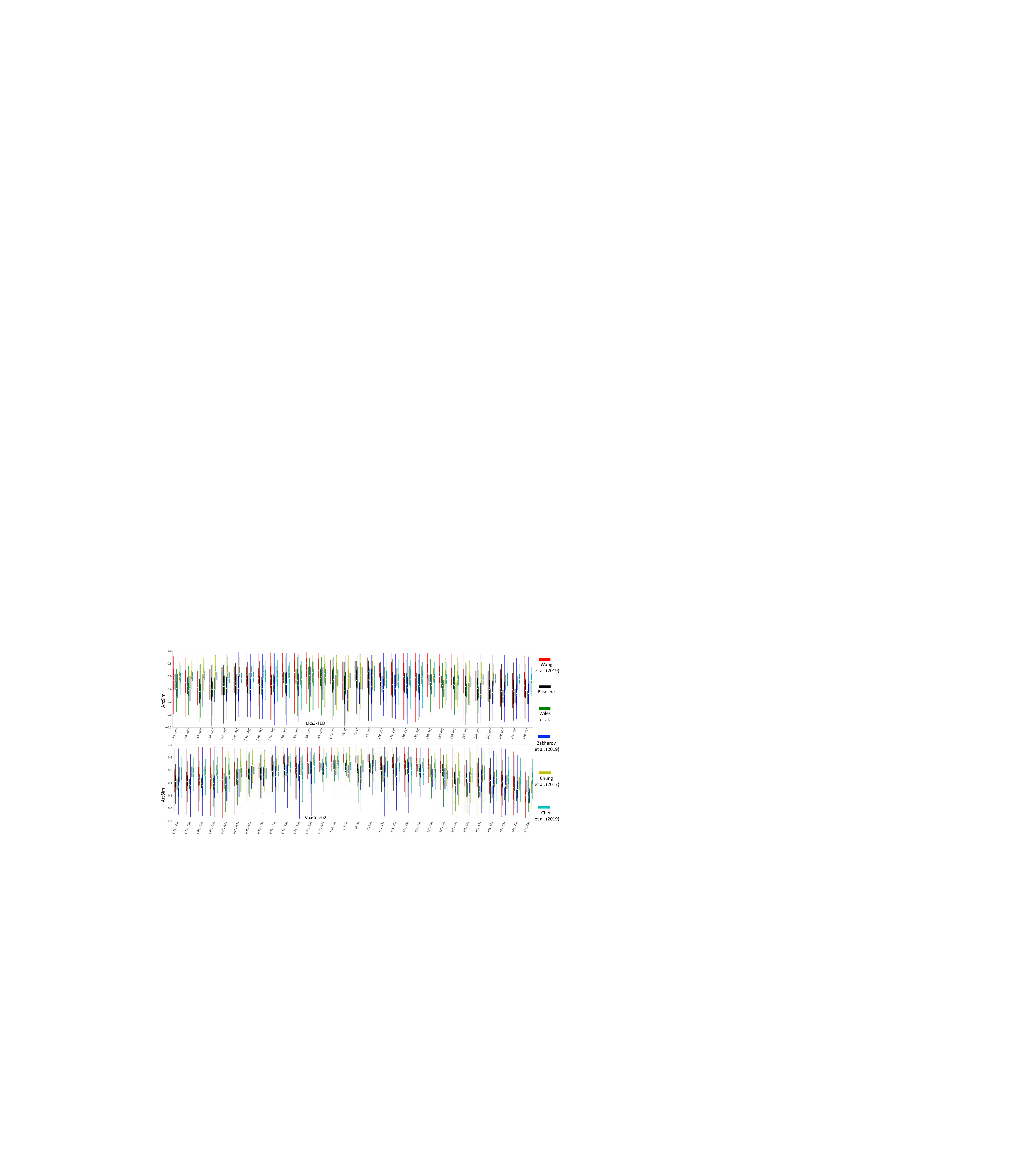}
\caption{Identity preserving performance versus head poses on LRS3-TED and VoxCeleb2 testing set. The X-axis and Y-axis are head pose degree, and ArcSim score. Right side shows the bar colors of different methods.}
\label{fig:csim_all}       
\end{figure*}

\section{Results and Discussion }
We survey the state-of-the-art identity-independent talking-head video generation approaches in Sect.~\ref{sec:generation}, and introduce benchmark datasets and metrics in Sect.~\ref{sec:dataset} and Sect.~\ref{sec:evaluation}.
In this section, we explicitly evaluate different approaches under various protocols to show the strengths and weaknesses of each model.

\subsection{Our Baseline}

To better understand what comprise a good talking-head generation model, we introduce a baseline model, which is equipped with
novel components proposed by previous works. 
We adopt the state-of-the-art video generation model proposed by Wang et al.~\citeyearpar{wang2019few} as the backbone and optimize the network structure to improve the performance 
. 
Specifically, in the image embedding module, Wang et al.~\citeyearpar{wang2019few} extract attention vectors directly from reference and target landmarks and apply them as weights to aggregate appearance patterns, with the purpose of collecting information from different degrees of talking head samples.
To better consider the connection between reference images and landmarks, we adopt Information Exchange module before the attention combination process. Moreover, we introduce the ConvGate module to extract additional attention vectors within reference images or landmarks, which could reduce the effect of noise and enhance common information among different samples that relate to important features such as identity.
Meanwhile, instead of using image matting function to combine warped images with raw output of the generator, we exploit a multi-branch non-linear combination module to compose them at a feature-level inside the generator.
Such method has the advantage of reducing artifacts caused by misalignment. The following experiment section will show the improvement that our novel modules can provide.

\subsection{Identity Preserving}
Evaluating the identity preserving performance in talking-head videos is challenging. 
Recall that in Sect.~\ref{subsec:eval_identity} we select ArcSim as the evaluation metric, which uses ArcFace to extract the identity embeddings and 
cosine similarity to measure the distance between the synthesized video frame and ground truth.
We average the similarity score over all the video frames.  Tab.~\ref{tab:overall} (ArcSim) columns show the identity preserving evaluation results on VoxCeleb, LRS3-TED, LRW, and GRID datasets. 
We can find that Chen et al.~\citeyearpar{chen2019hierarchical} and Jamaludin et al.~\citeyearpar{jamaludin2019you} achieve much better ArcSim scores than other methods. 
We argue that these two methods take advantage of omitting the head motion modeling and generate talking-head images with fixed head pose, which is same as the reference frame. 
In this way, they only need to synthesize the audio-visual correlated facial regions (e.g., lip region), and their results would be much stable than other methods since they do not need to ``imagine'' other new facial regions caused by head movements. 

  \begin{figure}[t]
  \includegraphics[width=0.98 \linewidth]{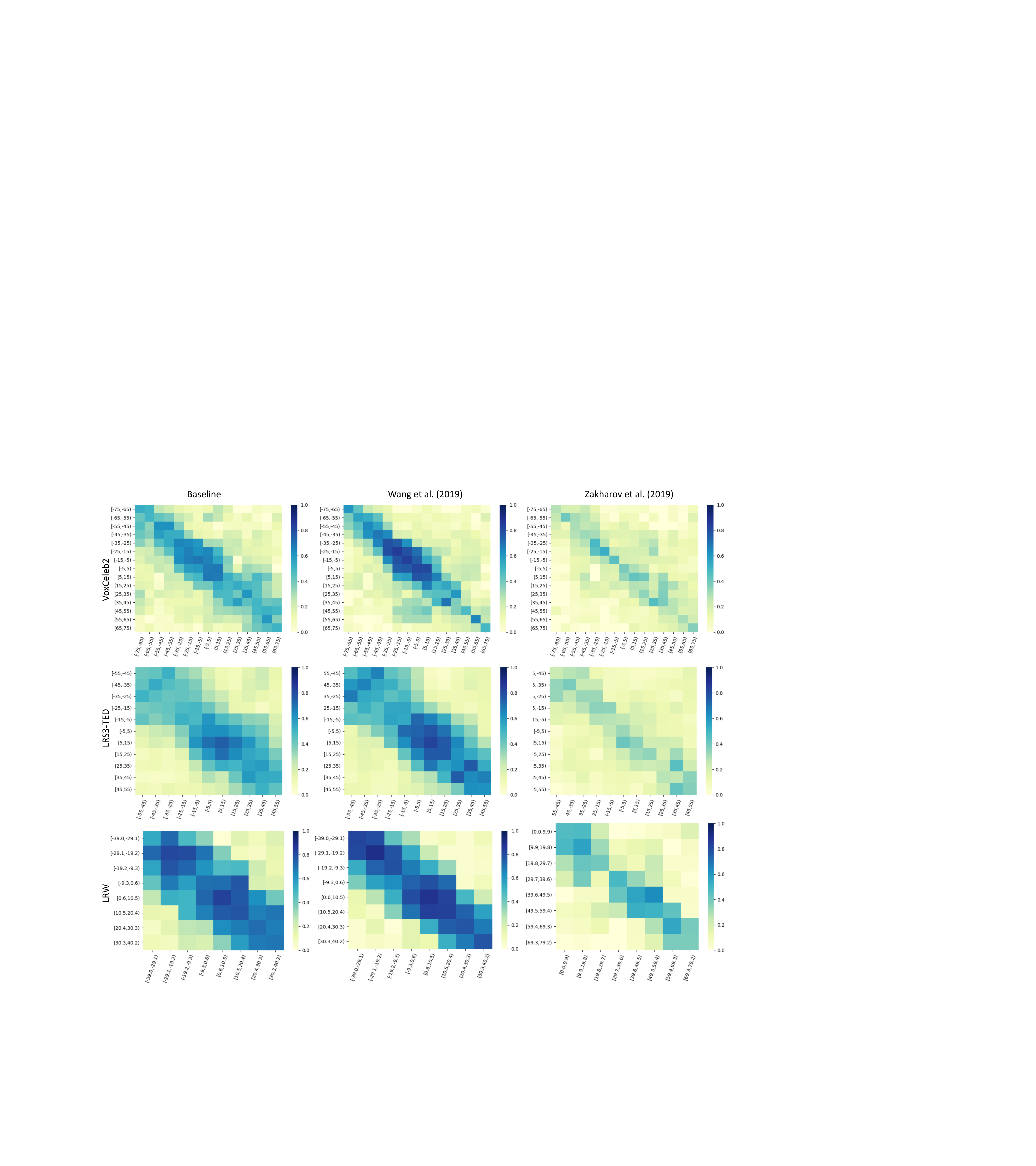}
\caption{The confusion matrix of ArcSim scores versus head poses of the reference video frame and target video frame. The first row shows the name of the methods, and the first column shows the dataset name.}
\label{fig:csim_matrix}       
\end{figure}

However, generating a profile face is more challenging than generating a frontal face (Tran et al.~\citeyear{tran2018representation}), and it is not fair to use the averaged ArcSim to represent the performance.
To explicitly study the correlation between the identity-preserving ability with the head pose of the target image frame, we show the results across different algorithms, and head poses in Fig.~\ref{fig:csim_all}.
We can find that all the methods can generate images with a better identity if the head pose of the target face is closer to the frontal.
This is an interesting phenomenon, which is not addressed in the previous talking-head generation papers.
To further dig the effects between head pose and identity-preserving performance, we also control the head pose of the reference images.
According to the confusion matrix (see Fig.~\ref{fig:csim_matrix}), it is obvious that the models can generate images with better identity if the head pose of the reference image and target image are close (the diagonal of each confusion matrix).
Meanwhile, this confusion matrix delivers another message that the ArcSim metric is robust to images with different head poses.

\begin{figure*}[t]
  \includegraphics[width=0.98 \linewidth]{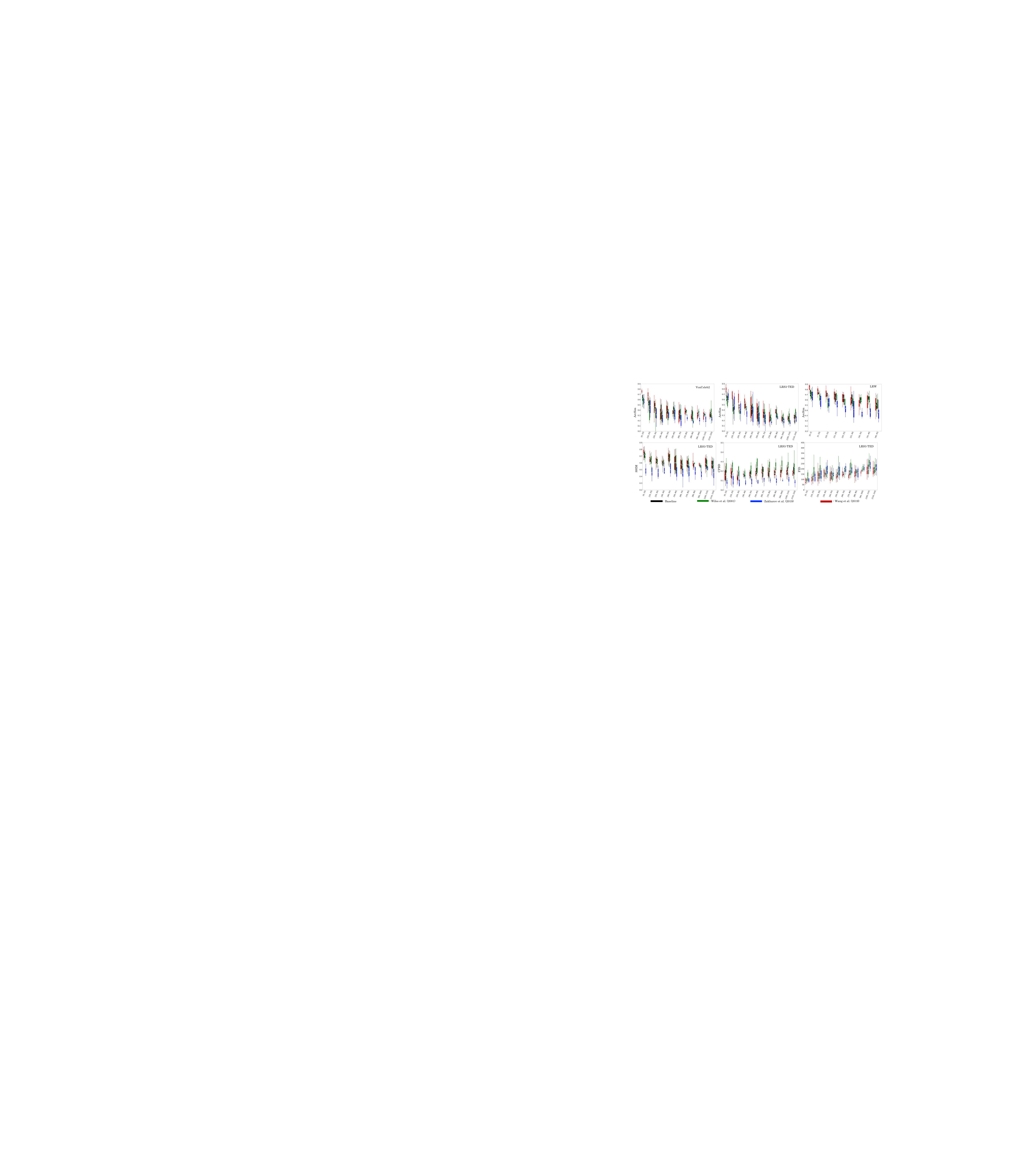}
\caption{Different evaluation metrics versus head motion score. Videos are divided into different bins grouped by the motion score. First row shows the ArcSim scores across different methods and datasets. The second row shows the visual quality evaluation results on LRS3-TED testing set across different methods. The X-axis is the degree bins of head motion. The last row shows the color bars of different methods.}
\label{fig:motion_all}       
\end{figure*}

After studying the relationship between identity-preserving ability and head pose, we can now reach an agreement that the averaged ArcSim score over the testing video samples can only partially demonstrate the identity-preserving performance since we do not know the head pose distribution.
To better utilize the ArcSim to evaluate video-level identity similarity, we plot the visual quality trend in a single video sampled from the VoxCeleb2 dataset. 
As shown in Fig.~\ref{fig:trend_pose}, as a
person moves head (\textcolor{black}{black line}), the ArcSim score (\textcolor{red}{red line}) changes significantly.
Meanwhile, we can find that the synthesized frames with the worst identity usually appear around the head pose boundaries.
For example, in Fig.~\ref{fig:trend_pose}, the frames with the lowest ArcSim score are frame $\#36$ and frame $\#96$, which are the frames with the rightmost
and leftmost head poses.
Inspired by this, in order to evaluate ArcSim in video-level fairly, we consider another property of talking-head---head motion.
We divide videos into different bins grouped by the motion score. The first row of Fig.~\ref{fig:motion_all} shows the ArcSim results across different methods and different datasets, where we can observe that all the methods achieve better ArcSim score on videos with smaller head motion degrees. 
This interesting finding can provide two suggestions to the future research: We should consider the balance of head motion when we compose the testing set for talking-head generation task; There is large improvement space for current talking-head video generation approaches, especially when generating videos with large head motions. 

\subsection{Visual Quality}
Recall the discussion in Sect.~\ref{subsec:visual_metrics}, we select SSIM and FID to measure the perceived error and quality of generated video frames, compared to the ground truth frames, which better mimics human perception.
CPBD is selected to determine blur based on the presence of edges in the synthesized video frames.
Tab.~\ref{tab:overall} shows the results across different datasets and methods. From CPBD, we observe that photo-metric $L1$ loss would decrease the sharpness of synthesized video frames (e.g., Jamaludin et al.~\citeyear{jamaludin2019you}; Chen et al.~\citeyear{chen2019hierarchical}). 
Since some methods omit the head motion modeling and generate static talking-head (Chen et al.~\citeyear{chen2019hierarchical}; Jamaludin et al.~\citeyear{jamaludin2019you}), the SSIM and FID scores of these methods are much lower than other methods considering the head movement. 

If we look at the SSIM (green line), and FID (yellow line) in Fig.~\ref{fig:trend_pose}, we can find that the visual quality of synthesized frames fluctuates
when there is an apparent head motion. Based on this observation, we group the videos by their head motion score and plot out the visual quality evaluation results versus head motion bins in the second row of Fig~\ref{fig:motion_all}. From the plot, we can obtain the similar findings as we observed in Fig.~\ref{fig:trend_pose}: all the methods have worse visual quality performance (SSIM and FID scores) on the videos in larger head motion bins, the sharpness of synthesized video does not effected by the head motion. 

\begin{figure}[t]
  \includegraphics[width=0.98 \linewidth]{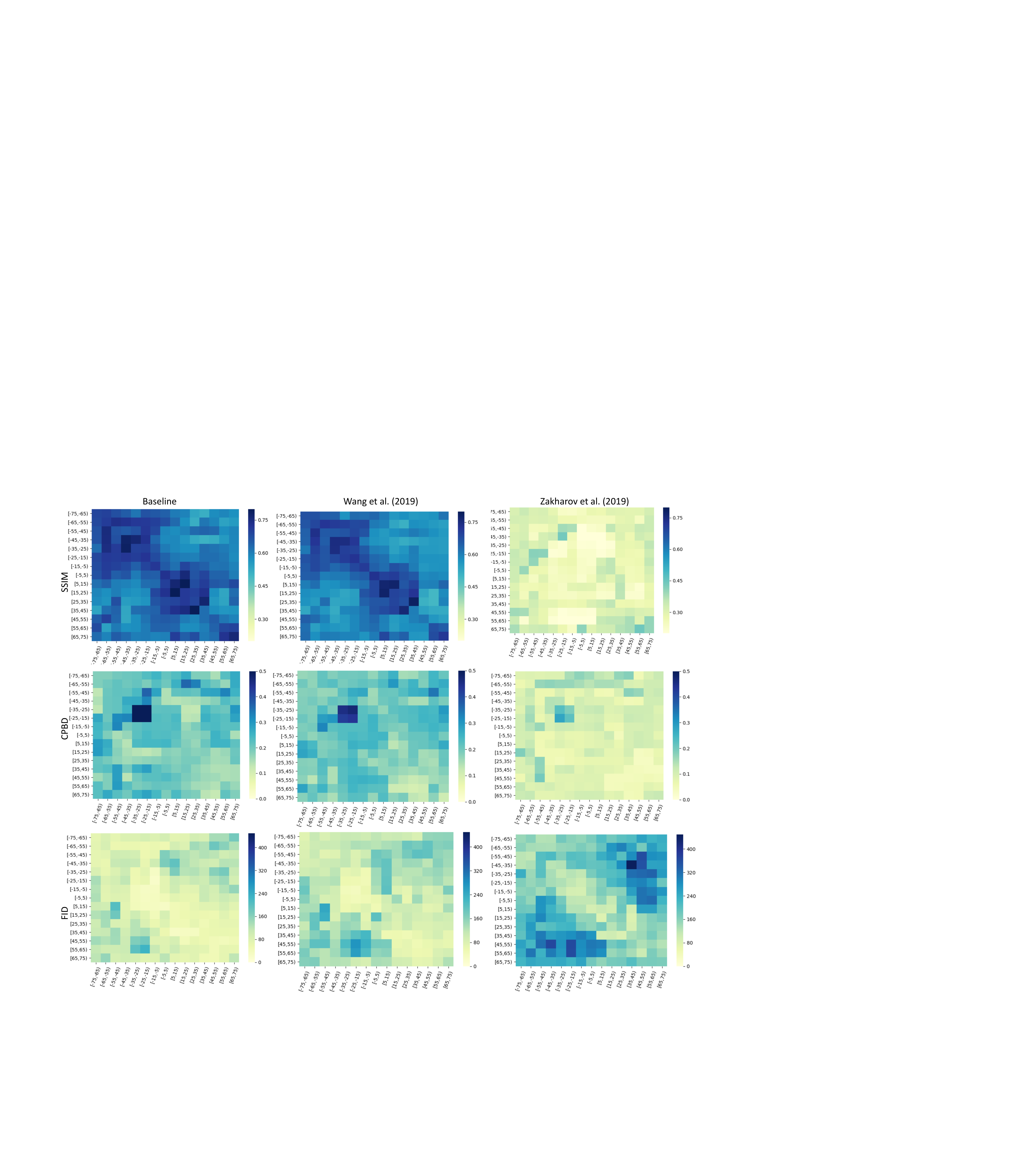}
\caption{The confusion matrix of visual quality results versus the head motion. The first row shows the name of methods, and the first column shows the name of the visual metrics. The X-axis and Y-axis are the head motions of reference frame and target frame.}
\label{fig:ssim_comfusion}       
\end{figure}

To investigate the correlation between the visual quality of synthesized video frames and the head pose, we plot the confusion matrix (Fig.~\ref{fig:ssim_comfusion}) for SSIM, CPBD, and FID with respective to the head pose bins of reference and target images on VoxCeleb2 dataset.
As shown in the second row of Fig.~\ref{fig:ssim_comfusion}, there is no clear correlation between the CPBD score and head poses, which is consistent with the definition of CPBD, since the sharpness is decided by algorithms (e.g., loss function, network structure) rather than data. 
For example, Wang et al.~\citeyearpar{wang2019few} and baseline achieve better performance than Zakharov et al.\citeyearpar{zakharov2019few} in terms of CPBD since their generator structure is much more complicated (e.g., SPADE blocks). 
If we look at the FID and SSIM (first and third row), it is obvious that all the methods can generate images with better visual quality if the head pose of the reference image and target image are close (the diagonal of each confusion matrix). 

\begin{table*}
\centering
\newcommand{\mr}[1]{\multirow{2}{*}{#1}}
\newcommand{\mrs}[2]{\multirow{2}{*}{\shortstack{#1\\#2}}}
\caption{Semantic-level video quality of different methods. The $L2$ row shows the $L2$ distance between features of the fake video and the paired real video.}
\resizebox{0.98\linewidth}{!}{
\newcolumntype{L}{>{\centering\arraybackslash}m{1.9cm}}
\newcolumntype{S}{>{\centering\arraybackslash}m{1.0cm}}
\newcolumntype{M}{>{\centering\arraybackslash}m{1.5cm}}
\begin{tabular}{S|S|M|S|M|M|L|S|M|S|M|M|L}
\toprule
\hline
&\multicolumn{6}{c|}{ LRS3-TED} &\multicolumn{6}{c}{ VoxCeleb2}\\
\hline
&\mrs{Real}{video}&\mrs{Chen et al.}{\citeyearpar{chen2019hierarchical}} &\mr{Baseline}&\mrs{Wang et al.}{\citeyearpar{wang2019few}}&\mrs{Wiles et al.}{\citeyearpar{x2face}}&\mrs{Zakharov et al.}{\citeyearpar{zakharov2019few}}&\mrs{Real}{Video}&\mrs{Chen et al.}{\citeyearpar{chen2019hierarchical}} &\mr{Baseline}&\mrs{Wang et al.}{\citeyearpar{wang2019few}}&\mrs{Wiles et al.}{\citeyearpar{x2face}}&\mrs{Zakharov et al.}{\citeyearpar{zakharov2019few}}\\
&&&&&&&&&&&\\
\hline
Top-1 &72.62\% & 1.99\% & \textbf{3.85\%} & 2.23\% & 1.93\% & 1.77\%  &42.46\% & 2.40\% & 
\textbf{3.04\%} & 1.64\% & 1.87\% & 1.99\%\\
Top-5 &87.98\% & 5.01\% & \textbf{11.05\%} & 6.19\% & 5.73\% & 5.48\%  &63.98\% &\textbf{8.12\%} &7.13\% &4.56\% &4.80\% &4.74\% \\
Top-10 &91.53\% & 8.19\% & \textbf{16.13\%} & 8.37\% & 8.62\% & 8.27\% &70.82\% &\textbf{10.76\%} &\textbf{10.76\%} &7.13\% &7.49\% &7.54\% \\
Top-20 &94.42\% & 12.99\% & \textbf{22.11\%} & 12.27\% & 13.18\% & 12.53 \% &78.30\%&15.80\%&\textbf{15.91\%}&11.23\%&11.70\%&11.05\%\\
LRSD &---&46.35 \% & \textbf{59.60\%} & 56.25\% & 52.95\% & 51.93\% &---&47.93\%&\textbf{62.56\%}&61.59\%&55.14\%&53.87\% \\ 
L2 &---&1.03&\textbf{0.89} &0.93 &0.96 &0.98 &---&1.02&\textbf{0.86}&0.87&0.94&0.96 \\
\hline 
\bottomrule
\end{tabular}
}
\label{tab:lipread}
\vspace{-5mm}
\end{table*}

\subsection{Semantic-level Lip Synchronization}

We use LRSD (Sect.~\ref{subsec:LRSD}) to evaluate the semantic-level quality of synthesized videos. Recall that our Lipreading network is trained on LRS3-TED training set, and works for videos outside the LRS3-TED dataset. Thus, we evaluate the semantic-level quality on videos from LRS3-TED and VoxCeleb2 testing set. Tab.~\ref{tab:lipread} shows the lipreading accuracy and LRSD scores. First, if we look at the lipreading accuracy on real videos from LRS3-TED and VoxCeleb2 testing set, we can find that our lipreading network achieves $72.62\%, 42.46\%$ top-1 accuracy on LRS3-TED and VoxCeleb2 testing sets, respectively, which demonstrates the generalizability of our lip-reading network. Since we want to compare the semantic-level quality of the synthesized videos by comparing the similarity distance between synthesized videos and real videos, we also care about the top-5, top-10, and top-20 accuracy. Then, we can look at the lipreading accuracy on synthesized videos. The incredibly low top-1 lipreading accuracy on the synthesized videos generated by all the methods indicate that these methods can not generate accurate lip movements for word-level lipreading classification. However, when we increase the tolerance to top-5, top-10, and top-20, the gap between different methods is growing, which means that those synthesized videos can still reflect its semantic-level meaning. Next, from the lipreading accuracy and LRSD scores, we can see that the LRSD performance are almost consistent with the lipreading accuracy (most of the methods with higher lipreading accuracy achieve better LRSD score). We also calculate the $L2$ distance of the lipreading features between synthesized video and paired ground truth, from which we can obtain similar conclusion. 

\begin{figure*}[t]
  \includegraphics[width=0.98 \linewidth]{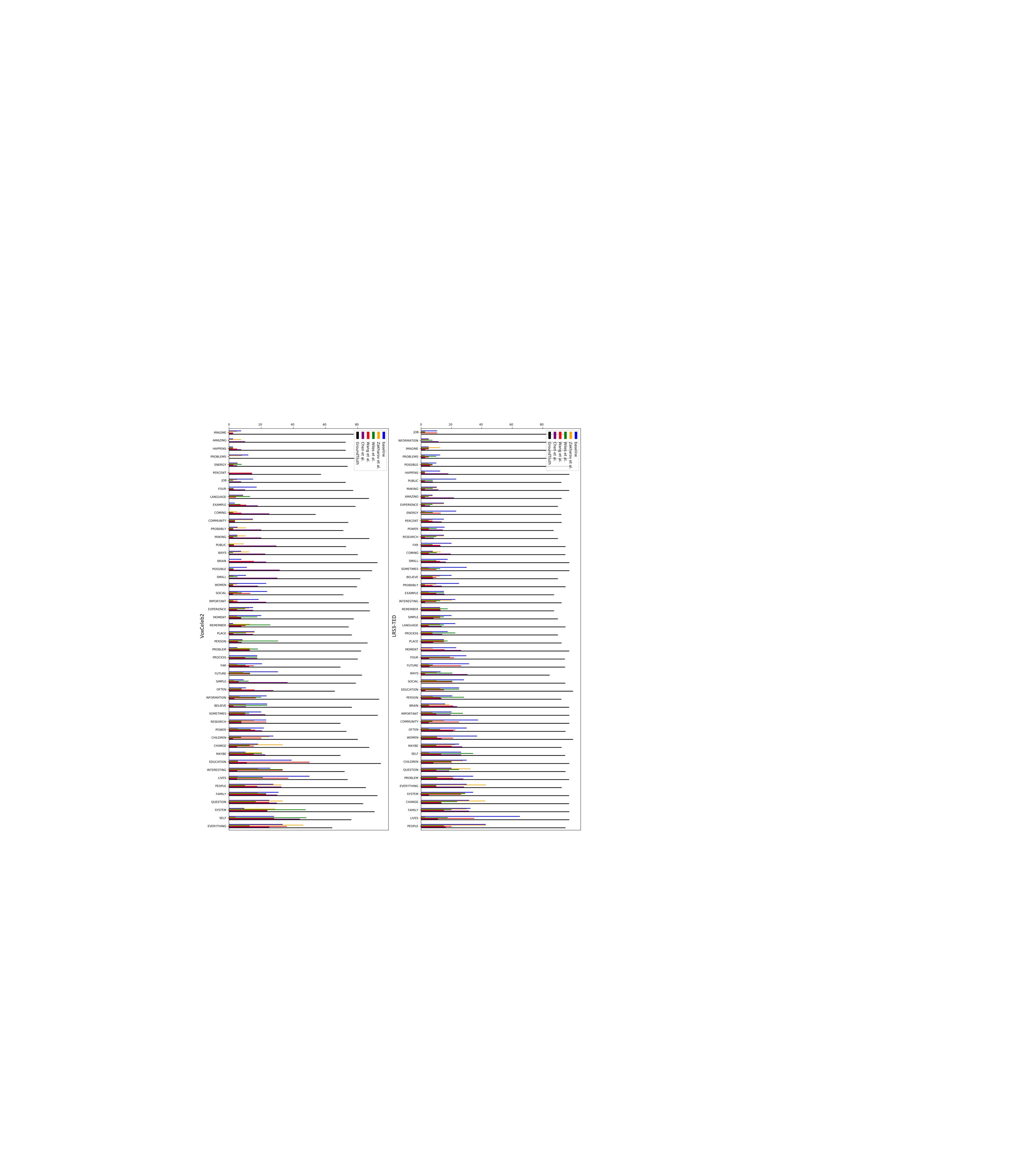}
\caption{The detailed top-20 lipreading accurate of synthesized videos and ground truth videos on VoxCeleb2 and LRS3-TED testing set. Please zoom in to look at the details.}
\label{fig:word_lip}       
\end{figure*}

In order to understand which words are more difficult to synthesize, we plot the lipreading accuracy of fake videos versus the word label in Fig.~\ref{fig:word_lip}. We can see that all the models do not perform well in synthesizing the lip motion for some words (e.g., `JOB', `POWER', 'IMAGINE', and `PUBLIC') in both VoxCeleb2 and LRS3-TED testing sets. There could be some common features in those words that current motion modeling methods can not handle, and this could be a good direction to boost the semantic-level performance of synthesized video. 

\subsection{Spontaneous Motion}
\label{subsec:spontMotionExp}

  \begin{figure}[t]
  \includegraphics[width=0.98 \linewidth]{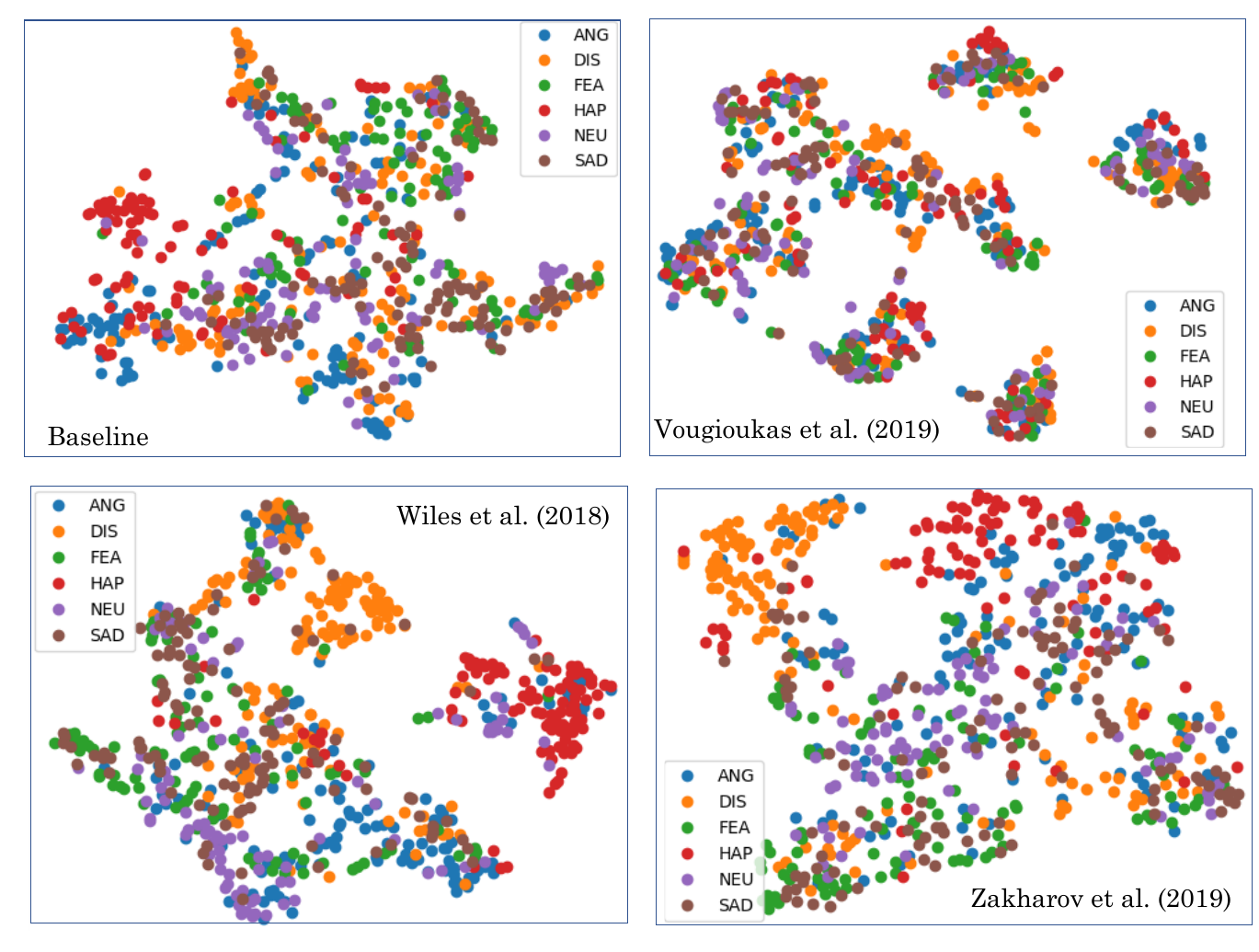}
\caption{The t-SNE plot of the ESD features of baseline, Vougioukas et al.~\citeyearpar{vougioukas2019realistic}. Wiles et al.~\citeyearpar{x2face}, and Zakharov et al.~\citeyearpar{zakharov2019few}. }
\label{fig:emotion_tsne}       
\end{figure}

\begin{table*}[t]
	\centering
	\caption{The ESD score across over different methods on CREMA-D testing set. ESD measures the cosine similarity between the emotion features extracted from synthesized video and ground truth video. We bold the leading score.}
	\begin{tabular}{ ccccc}
		\toprule
		\hline
Method & Baseline& Zakharov et al.~\citeyearpar{zakharov2019few} & Wiles et al.~\citeyearpar{x2face}   &Vougioukas et al.~\citeyearpar{vougioukas2019realistic} \\  \hline 
ESD $\uparrow$ \  & 0.467    & 0.391& \textbf{0.655} & 0.2665 \\       
 \bottomrule
	\end{tabular}
	\label{Table:esd}
\end{table*}

\begin{table}
	\centering
	\caption{The BSD score of different methods on Grid testing set. BSD score measures cosine similarity between blink features extracted from synthesized video and ground truth video. As mentioned in sec~\ref{subsec:sec4_motion}, we sample several slices from each video for related blink features.}
	\begin{tabular}{c|c}
		\toprule
		\hline
Method & BSD $\uparrow$ \\\hline
Baseline & 0.965 \\
Zakharov et al.~\citeyearpar{zakharov2019few} & 0.935 \\
Vougioukas et al.~\citeyearpar{vougioukas2019realistic}  & 0.919 \\
Chen et al.~\citeyearpar{chen2018lip} & 0.907 \\
Jamaludin et al.~\citeyearpar{jamaludin2019you} & 0.807 \\
Wang et al.~\citeyearpar{wang2019few} & 0.957 \\
Wiles et al.~\citeyearpar{x2face} & \textbf{0.979}\\
 \bottomrule
	\end{tabular}
	\label{Table:bsd}
\end{table}

\begin{table}[t]
\scriptsize
    \centering
    \caption{The results across different methods on ObamaSet. Specifically, we report ArcSim, SSIM, CPBD and FID of methods including baseline, Wang et al.~\citeyearpar{wang2019few}, Zakharov et al.~\citeyearpar{zakharov2019few} and Wiles et al.~\citeyearpar{x2face}.}
    \begin{tabular}{c|c|c|c|c}
        \toprule
        \hline
        Method & ArcSim & SSIM & CPBD & FID  \\\hline
        Baseline & \textbf{0.76} & 0.42 & \textbf{0.32} & 132.03 \\
        Wang et al.~\citeyearpar{wang2019few} & 0.49 & \textbf{0.52} & 0.22 & \textbf{83.61} \\
        Zakharov et al.~\citeyearpar{zakharov2019few} & 0.61 & 0.49 & 0.13 & 87.77 \\
        Wiles et al.~\citeyearpar{x2face} & 0.53 & \textbf{0.52} & 0.10 & 129.71 \\\hline
    \end{tabular}
    \label{tab:my_label}
\end{table}

We evaluate the emotional state of synthesized talking-head videos using the proposed Emotion Similarity Distance (ESD, see Sect.~\ref{subsec:sec4_motion}) on CREMA-D testing set. Fig.~\ref{fig:emotion_cls} first row shows the emotion classification accuracy on different types of videos. The purple, blue, red, orange, and green bars are the results on real videos, fake videos synthesized by Vougioukas et al.~\citeyearpar{vougioukas2019realistic}, fake videos from Wiles et al.~\citeyearpar{x2face}, fake videos from Zakharov et al.~\citeyear{zakharov2019few}, and fake videos synthesized by our baseline model, respectively. Fig.~\ref{fig:emotion_tsne} shows the t-SNE plot of different ESD features on CREMA-D testing set, from where we observe that the group boundaries of ESD feature extracted from baseline method, Wiles et al.~\citeyearpar{x2face}, and Zakharov et al.~\citeyearpar{zakharov2019few} are more clear than the ESD feature extracted from synthesized videos produced by Vougioukas et al.~\citeyearpar{vougioukas2019realistic}. The t-SNE visualization is consistent with the classification results in the first row of Fig.~\ref{fig:emotion_tsne}, where the emotion classifier achieves lowest accuracy on synthesized videos produced by Vougioukas et al.~\citeyearpar{vougioukas2019realistic}.  Tab.~\ref{Table:esd} shows the quantitative result of ESD, from where we can find that the emotional feature extracted from Wiles et al.~\citeyearpar{x2face} is closest to the feature extracted from ground truth comparing to other methods. This is consistent with the emotion classification accuracy shown in Fig.~\ref{fig:emotion_cls} second row, where the synthesized videos produced by Wiles et al.~\citeyearpar{x2face} achieves highest classification accuracy ($45.3\%$). In summary, the results shown in Fig.~\ref{fig:emotion_cls}, Fig.~\ref{fig:emotion_tsne}, and Tab.~\ref{Table:esd} demonstrate that our ESD is a well-characterized perceptual similarity measure that aims to assess the emotional expression ability of synthesized videos.

We also investigate the long-term motion modeling on ObamaSet (Tab.~\ref{tab:my_label}) since it contains more than 14 hours footage videos for the single subject. While all the selected methods can learn some general motions from the training data, there is no direct correlation between the synthesized motion with the input condition (e.g., audio, facial landmarks). We attribute this to the lack of specific module in those methods to model the individual's motion, thus leading to some random repeated motions in the synthesized videos.

We also analyze blink motion quality of talking-head videos generated by different methods. In Fig.~\ref{fig:blink_accuracy}, we use histogram to represent accuracy over blink classification on GRID's testing set where pink, cyan, magenta, gray, green, sky blue, orange and olive bars refer to Chen et al.~\citeyearpar{chen2018lip}, Jamaludin et al.~\citeyearpar{jamaludin2019you}, Vougioukas et al.~\citeyear{vougioukas2019realistic}, baseline,  Zakharov et al.~\citeyear{zakharov2019few}, Wiles et al.~\citeyear{x2face}, Wang et al.~\citeyear{wang2019few} and ground truth video specifically. We find that the baseline achieves a high performance on blink motion synthesizing, whose accuracy is approximately equal to ground truth blink motion. Moreover, comparing between landmark driven methods (baseline,  Zakharov et al.~\citeyear{zakharov2019few}, Wiles et al.~\citeyear{x2face}), Wang et al.~\citeyear{wang2019few}) and audio driven methods (Chen et al.~\citeyearpar{chen2018lip}, Jamaludin et al.~\citeyearpar{jamaludin2019you}, Vougioukas et al.~\citeyear{vougioukas2019realistic}), we can find that the former tend to have higher accuracy on overall testing set. This result is reasonable, as landmark provides direct clue for the motion of eyes. However, noise also exists in landmarks. We observe that even for some frames representing no changes in eyes region, obvious different open rates are represented in related landmarks. 
Misled by these noise, landmark driven methods may also result in wrong blink motion (e.g., Zakharov et al.~\citeyear{zakharov2019few} in fig~\ref{fig:blink_accuracy}). Furthermore, we also calculate BSD score for each method and show them in Tab.~\ref{Table:bsd}. Similarly, we can find that landmark driven methods always synthesize more similar blinking motion as ground truth talking head and result in higher BSD score. We also plot t-SNE figure for synthesized videos for each method. Also as expected, landmark driven methods show more obvious clustering ability for different blink class compared with audio driven methods. Observing among the four landmark driven methods, Zakharov et al.~\citeyearpar{zakharov2019few} shows bottleneck in separating blink motion video and non-blink motion video, which is the same as result of classification accuracy and BSD. All these experiments show the ability of blink model and BSD to evaluate blinking performance of synthesized videos and the results are reasonable. Moreover, this may also lead us to think about the robustness of GAN over noise in landmarks.

  \begin{figure}[t]
  \includegraphics[width=0.98 \linewidth]{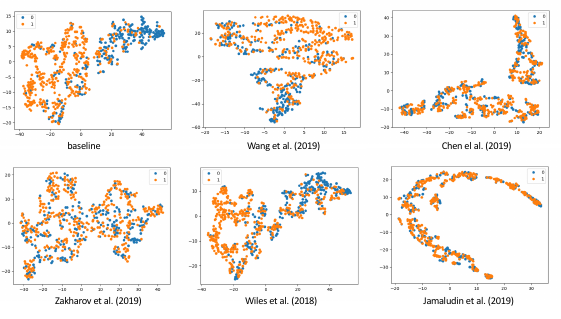}
\caption{t-SNE plot for blink features of videos synthesized by different methods on GRID's testing set. For each figure, points with label 1 refer to slices with ground truth blink motion; points with label 0 refer to slices with ground truth non-blink motion.}
\label{fig:blink_tsne}       
\end{figure}

\section{Conclusion}
Talking-head generation is an important and challenging
problem in computer vision and has received considerable attention. 
Thanks to remarkable developments in GAN techniques, the field of talking-head generation has dramatically
evolved. 
As a comprehensive survey on talking-head generation task and its evaluation metrics, this paper has highlighted recent achievements, provided well-defined standards for comprising a good talking-head video, summarized existing popular datasets and evaluation criteria, conducted detailed comparative empirical and analytical studies of available measure, and benchmarked models under the same conditions using more than one measure. 
Meanwhile, we introduced three perceptually meaningful metrics that assess the emotional expression, semantic-level lip synchronization, and blink motion of a synthesized video. The proposed metrics agree with human perceptual judgment, and have low sample and computational complexity. The performance of talking-head generation will continue to improve as various structures are proposed. In the mean time, seeking appropriate measures for this task continues to be an important open problem, not only for fair model comparison but also for understanding, improving, and developing the talking-head animation models.

\vspace{5mm}
\noindent \textbf{Acknowledgments.} \quad This work was supported in part by NSF 1741472, 1813709, and 1909912.
The article solely reflects the opinions and conclusions of its authors but not the funding agents.

\newpage
\bibliographystyle{spbasic}      
\bibliography{cite}   

\begin{thebibliography}{71}
\providecommand{\natexlab}[1]{#1}
\providecommand{\url}[1]{{#1}}
\providecommand{\urlprefix}{URL }
\expandafter\ifx\csname urlstyle\endcsname\relax
  \providecommand{\doi}[1]{DOI~\discretionary{}{}{}#1}\else
  \providecommand{\doi}{DOI~\discretionary{}{}{}\begingroup
  \urlstyle{rm}\Url}\fi
\providecommand{\eprint}[2][]{\url{#2}}

\bibitem[{Afouras et~al.(2018{\natexlab{a}})Afouras, Chung, Senior, Vinyals,
  and Zisserman}]{afouras2018deep}
Afouras T, Chung JS, Senior A, Vinyals O, Zisserman A (2018{\natexlab{a}}) Deep
  audio-visual speech recognition. IEEE transactions on pattern analysis and
  machine intelligence

\bibitem[{Afouras et~al.(2018{\natexlab{b}})Afouras, Chung, and
  Zisserman}]{lrs3}
Afouras T, Chung JS, Zisserman A (2018{\natexlab{b}}) Lrs3-ted: a large-scale
  dataset for visual speech recognition. arXiv preprint arXiv:180900496

\bibitem[{Assael et~al.(2016)Assael, Shillingford, Whiteson, and
  De~Freitas}]{assael2016lipnet}
Assael YM, Shillingford B, Whiteson S, De~Freitas N (2016) Lipnet: End-to-end
  sentence-level lipreading. arXiv preprint arXiv:161101599

\bibitem[{Averbuch-Elor et~al.(2017)Averbuch-Elor, Cohen-Or, Kopf, and
  Cohen}]{averbuch2017bringing}
Averbuch-Elor H, Cohen-Or D, Kopf J, Cohen MF (2017) Bringing portraits to
  life. ACM Transactions on Graphics (TOG) 36(6):196

\bibitem[{Bregler et~al.(1997)Bregler, Covell, and Slaney}]{bregler1997video}
Bregler C, Covell M, Slaney M (1997) Video rewrite: Driving visual speech with
  audio. In: Proceedings of the 24th annual conference on Computer graphics and
  interactive techniques, pp 353--360

\bibitem[{Bulat and Tzimiropoulos(2017)}]{bulat2017far}
Bulat A, Tzimiropoulos G (2017) How far are we from solving the 2d \& 3d face
  alignment problem?(and a dataset of 230,000 3d facial landmarks). In:
  Proceedings of the IEEE International Conference on Computer Vision, pp
  1021--1030

\bibitem[{Busso et~al.(2016)Busso, Parthasarathy, Burmania, AbdelWahab,
  Sadoughi, and Provost}]{busso2016msp}
Busso C, Parthasarathy S, Burmania A, AbdelWahab M, Sadoughi N, Provost EM
  (2016) Msp-improv: An acted corpus of dyadic interactions to study emotion
  perception. IEEE Transactions on Affective Computing 8(1):67--80

\bibitem[{Cao et~al.(2014)Cao, Cooper, Keutmann, Gur, Nenkova, and
  Verma}]{crema}
Cao H, Cooper DG, Keutmann MK, Gur RC, Nenkova A, Verma R (2014) Crema-d:
  Crowd-sourced emotional multimodal actors dataset. IEEE transactions on
  affective computing 5(4):377--390

\bibitem[{Cassell et~al.(1999)Cassell, McNeill, and McCullough}]{cassel1999}
Cassell J, McNeill D, McCullough KE (1999) Speech-gesture mismatches: Evidence
  for one underlying representation of linguistic and nonlinguistic
  information. Pragmatics \& cognition 7(1):1--34

\bibitem[{Chan et~al.(2019)Chan, Ginosar, Zhou, and Efros}]{chan2019everybody}
Chan C, Ginosar S, Zhou T, Efros AA (2019) Everybody dance now. In: Proceedings
  of the IEEE International Conference on Computer Vision, pp 5933--5942

\bibitem[{Chang and Ezzat(2005)}]{chang2005transferable}
Chang YJ, Ezzat T (2005) Transferable videorealistic speech animation. In:
  Proceedings of the 2005 ACM SIGGRAPH/Eurographics symposium on Computer
  animation, pp 143--151

\bibitem[{Chechik et~al.(2010)Chechik, Sharma, Shalit, and
  Bengio}]{chechik2010large}
Chechik G, Sharma V, Shalit U, Bengio S (2010) Large scale online learning of
  image similarity through ranking. Journal of Machine Learning Research
  11(Mar):1109--1135

\bibitem[{Chen et~al.(2018)Chen, Li, K~Maddox, Duan, and Xu}]{chen2018lip}
Chen L, Li Z, K~Maddox R, Duan Z, Xu C (2018) Lip movements generation at a
  glance. In: Proceedings of the European Conference on Computer Vision (ECCV),
  pp 520--535

\bibitem[{Chen et~al.(2019)Chen, Maddox, Duan, and Xu}]{chen2019hierarchical}
Chen L, Maddox RK, Duan Z, Xu C (2019) Hierarchical cross-modal talking face
  generation with dynamic pixel-wise loss. In: Proceedings of the IEEE
  Conference on Computer Vision and Pattern Recognition, pp 7832--7841

\bibitem[{Chung and Zisserman(2016{\natexlab{a}})}]{chung2016lip}
Chung JS, Zisserman A (2016{\natexlab{a}}) Lip reading in the wild. In: Asian
  Conference on Computer Vision, Springer, pp 87--103

\bibitem[{Chung and Zisserman(2016{\natexlab{b}})}]{chung2016out}
Chung JS, Zisserman A (2016{\natexlab{b}}) Out of time: automated lip sync in
  the wild. In: Asian conference on computer vision, Springer, pp 251--263

\bibitem[{Chung and Zisserman(2017)}]{Chung17a}
Chung JS, Zisserman A (2017) Lip reading in profile. In: British Machine Vision
  Conference

\bibitem[{Chung et~al.(2018)Chung, Nagrani, and Zisserman}]{voxceleb2}
Chung JS, Nagrani A, Zisserman A (2018) Voxceleb2: Deep speaker recognition.
  arXiv preprint arXiv:180605622

\bibitem[{Cooke et~al.(2006)Cooke, Barker, Cunningham, and Shao}]{grid}
Cooke M, Barker J, Cunningham S, Shao X (2006) An audio-visual corpus for
  speech perception and automatic speech recognition. The Journal of the
  Acoustical Society of America 120(5):2421--2424

\bibitem[{Czyzewski et~al.(2017)Czyzewski, Kostek, Bratoszewski, Kotus, and
  Szykulski}]{czyzewski2017audio}
Czyzewski A, Kostek B, Bratoszewski P, Kotus J, Szykulski M (2017) An
  audio-visual corpus for multimodal automatic speech recognition. Journal of
  Intelligent Information Systems 49(2):167--192

\bibitem[{Deng et~al.(2019)Deng, Guo, Xue, and Zafeiriou}]{deng2019arcface}
Deng J, Guo J, Xue N, Zafeiriou S (2019) Arcface: Additive angular margin loss
  for deep face recognition. In: Proceedings of the IEEE Conference on Computer
  Vision and Pattern Recognition, pp 4690--4699

\bibitem[{Essenwanger(1986)}]{essenwanger1986elements}
Essenwanger O (1986) Elements of Statistical Analysis. General climatology,
  Elsevier, \urlprefix\url{https://books.google.com/books?id=P6Y4vgAACAAJ}

\bibitem[{Fan et~al.(2015)Fan, Wang, Soong, and Xie}]{fan2015photo}
Fan B, Wang L, Soong FK, Xie L (2015) Photo-real talking head with deep
  bidirectional lstm. In: 2015 IEEE International Conference on Acoustics,
  Speech and Signal Processing (ICASSP), IEEE, pp 4884--4888

\bibitem[{Fried et~al.(2019)Fried, Tewari, Zollh{\"o}fer, Finkelstein,
  Shechtman, Goldman, Genova, Jin, Theobalt, and Agrawala}]{fried2019text}
Fried O, Tewari A, Zollh{\"o}fer M, Finkelstein A, Shechtman E, Goldman DB,
  Genova K, Jin Z, Theobalt C, Agrawala M (2019) Text-based editing of
  talking-head video. ACM Transactions on Graphics (TOG) 38(4):1--14

\bibitem[{Garrido et~al.(2015)Garrido, Valgaerts, Sarmadi, Steiner, Varanasi,
  Perez, and Theobalt}]{garrido2015vdub}
Garrido P, Valgaerts L, Sarmadi H, Steiner I, Varanasi K, Perez P, Theobalt C
  (2015) Vdub: Modifying face video of actors for plausible visual alignment to
  a dubbed audio track. In: Computer graphics forum, Wiley Online Library,
  vol~34, pp 193--204

\bibitem[{Ginosar et~al.(2019)Ginosar, Bar, Kohavi, Chan, Owens, and
  Malik}]{ginosar2019learning}
Ginosar S, Bar A, Kohavi G, Chan C, Owens A, Malik J (2019) Learning individual
  styles of conversational gesture. In: Proceedings of the IEEE Conference on
  Computer Vision and Pattern Recognition, pp 3497--3506

\bibitem[{Glowinski et~al.(2011)Glowinski, Dael, Camurri, Volpe, Mortillaro,
  and Scherer}]{glowinski2011toward}
Glowinski D, Dael N, Camurri A, Volpe G, Mortillaro M, Scherer K (2011) Toward
  a minimal representation of affective gestures. IEEE Transactions on
  Affective Computing 2(2):106--118

\bibitem[{Graves et~al.(2006)Graves, Fern{\'a}ndez, Gomez, and
  Schmidhuber}]{graves2006connectionist}
Graves A, Fern{\'a}ndez S, Gomez F, Schmidhuber J (2006) Connectionist temporal
  classification: labelling unsegmented sequence data with recurrent neural
  networks. In: Proceedings of the 23rd international conference on Machine
  learning, pp 369--376

\bibitem[{Gu et~al.(2019)Gu, Zhou, and Huang}]{gu2019flnet}
Gu K, Zhou Y, Huang T (2019) Flnet: Landmark driven fetching and learning
  network for faithful talking facial animation synthesis. arXiv preprint
  arXiv:191109224

\bibitem[{Harte and Gillen(2015)}]{tcd}
Harte N, Gillen E (2015) Tcd-timit: An audio-visual corpus of continuous
  speech. IEEE Transactions on Multimedia 17(5):603--615

\bibitem[{Heusel et~al.(2017)Heusel, Ramsauer, Unterthiner, Nessler, and
  Hochreiter}]{heusel2017gans}
Heusel M, Ramsauer H, Unterthiner T, Nessler B, Hochreiter S (2017) Gans
  trained by a two time-scale update rule converge to a local nash equilibrium.
  In: Advances in neural information processing systems, pp 6626--6637

\bibitem[{Jamaludin et~al.(2019)Jamaludin, Chung, and
  Zisserman}]{jamaludin2019you}
Jamaludin A, Chung JS, Zisserman A (2019) You said that?: Synthesising talking
  faces from audio. International Journal of Computer Vision
  127(11-12):1767--1779

\bibitem[{Jia et~al.(2014)Jia, Wu, Zhang, Meng, and Cai}]{jia2014head}
Jia J, Wu Z, Zhang S, Meng HM, Cai L (2014) Head and facial gestures synthesis
  using pad model for an expressive talking avatar. Multimedia Tools and
  Applications 73(1):439--461

\bibitem[{Karras et~al.(2017)Karras, Aila, Laine, Herva, and
  Lehtinen}]{karras2017audio}
Karras T, Aila T, Laine S, Herva A, Lehtinen J (2017) Audio-driven facial
  animation by joint end-to-end learning of pose and emotion. ACM Transactions
  on Graphics (TOG) 36(4):1--12

\bibitem[{Kim et~al.(2018)Kim, Garrido, Tewari, Xu, Thies, Nie{\ss}ner,
  P{\'e}rez, Richardt, Zollh{\"o}fer, and Theobalt}]{kim2018deep}
Kim H, Garrido P, Tewari A, Xu W, Thies J, Nie{\ss}ner M, P{\'e}rez P, Richardt
  C, Zollh{\"o}fer M, Theobalt C (2018) Deep video portraits. ACM Transactions
  on Graphics (TOG) 37(4):1--14

\bibitem[{Liu and Ostermann(2011)}]{liu2011realistic}
Liu K, Ostermann J (2011) Realistic facial expression synthesis for an
  image-based talking head. In: 2011 IEEE International Conference on
  Multimedia and Expo, IEEE, pp 1--6

\bibitem[{Liu et~al.(2019)Liu, Huang, Mallya, Karras, Aila, Lehtinen, and
  Kautz}]{liu2019few}
Liu MY, Huang X, Mallya A, Karras T, Aila T, Lehtinen J, Kautz J (2019)
  Few-shot unsupervised image-to-image translation. In: Proceedings of the IEEE
  International Conference on Computer Vision, pp 10551--10560

\bibitem[{Liu et~al.(2018)Liu, Luo, Lian, and Gao}]{liu2018future}
Liu W, Luo W, Lian D, Gao S (2018) Future frame prediction for anomaly
  detection--a new baseline. In: Proceedings of the IEEE Conference on Computer
  Vision and Pattern Recognition, pp 6536--6545

\bibitem[{Livingstone and Russo(2018)}]{ryerson}
Livingstone SR, Russo FA (2018) The ryerson audio-visual database of emotional
  speech and song (ravdess): A dynamic, multimodal set of facial and vocal
  expressions in north american english. PloS one 13(5)

\bibitem[{Maaten and Hinton(2008)}]{maaten2008visualizing}
Maaten Lvd, Hinton G (2008) Visualizing data using t-sne. Journal of machine
  learning research 9(Nov):2579--2605

\bibitem[{Minderer et~al.(2019)Minderer, Sun, Villegas, Cole, Murphy, and
  Lee}]{minderer2019unsupervised}
Minderer M, Sun C, Villegas R, Cole F, Murphy KP, Lee H (2019) Unsupervised
  learning of object structure and dynamics from videos. In: Advances in Neural
  Information Processing Systems, pp 92--102

\bibitem[{Nagrani et~al.(2017)Nagrani, Chung, and Zisserman}]{voxceleb1}
Nagrani A, Chung JS, Zisserman A (2017) Voxceleb: a large-scale speaker
  identification dataset. arXiv preprint arXiv:170608612

\bibitem[{Narvekar and Karam(2009)}]{narvekar2009no}
Narvekar ND, Karam LJ (2009) A no-reference perceptual image sharpness metric
  based on a cumulative probability of blur detection. In: 2009 International
  Workshop on Quality of Multimedia Experience, IEEE, pp 87--91

\bibitem[{{\"O}hman and Salvi(1999)}]{ohman1999using}
{\"O}hman T, Salvi G (1999) Using hmms and anns for mapping acoustic to visual
  speech. TMH-QPSR 40(1-2):45--50

\bibitem[{Park et~al.(2019)Park, Liu, Wang, and Zhu}]{spade}
Park T, Liu MY, Wang TC, Zhu JY (2019) Semantic image synthesis with
  spatially-adaptive normalization. In: Proceedings of the IEEE Conference on
  Computer Vision and Pattern Recognition, pp 2337--2346

\bibitem[{Parkhi et~al.(2015)Parkhi, Vedaldi, and Zisserman}]{parkhi2015deep}
Parkhi OM, Vedaldi A, Zisserman A (2015) Deep face recognition. In:
  Xianghua~Xie MWJ, Tam GKL (eds) Proceedings of the British Machine Vision
  Conference (BMVC), BMVA Press, pp 41.1--41.12, \doi{10.5244/C.29.41},
  \urlprefix\url{https://dx.doi.org/10.5244/C.29.41}

\bibitem[{Poria et~al.(2018)Poria, Hazarika, Majumder, Naik, Cambria, and
  Mihalcea}]{MELD2018}
Poria S, Hazarika D, Majumder N, Naik G, Cambria E, Mihalcea R (2018) Meld: A
  multimodal multi-party dataset for emotion recognition in conversations.
  arXiv preprint arXiv:181002508

\bibitem[{Pumarola et~al.(2019)Pumarola, Agudo, Martinez, Sanfeliu, and
  Moreno-Noguer}]{ganimation}
Pumarola A, Agudo A, Martinez AM, Sanfeliu A, Moreno-Noguer F (2019)
  Ganimation: One-shot anatomically consistent facial animation. International
  Journal of Computer Vision pp 1--16

\bibitem[{Rossler et~al.(2019)Rossler, Cozzolino, Verdoliva, Riess, Thies, and
  Nie{\ss}ner}]{facefor}
Rossler A, Cozzolino D, Verdoliva L, Riess C, Thies J, Nie{\ss}ner M (2019)
  Faceforensics++: Learning to detect manipulated facial images. In:
  Proceedings of the IEEE International Conference on Computer Vision, pp 1--11

\bibitem[{Salimans et~al.(2016)Salimans, Goodfellow, Zaremba, Cheung, Radford,
  and Chen}]{salimans2016improved}
Salimans T, Goodfellow I, Zaremba W, Cheung V, Radford A, Chen X (2016)
  Improved techniques for training gans. In: Advances in neural information
  processing systems, pp 2234--2242

\bibitem[{Siarohin et~al.(2019{\natexlab{a}})Siarohin, Lathuili{\`e}re,
  Sangineto, and Sebe}]{siarohin2019appearance}
Siarohin A, Lathuili{\`e}re S, Sangineto E, Sebe N (2019{\natexlab{a}})
  Appearance and pose-conditioned human image generation using deformable gans.
  IEEE Transactions on Pattern Analysis and Machine Intelligence

\bibitem[{Siarohin et~al.(2019{\natexlab{b}})Siarohin, Lathuilière, Tulyakov,
  Ricci, and Sebe}]{NIPS2019_8935}
Siarohin A, Lathuilière S, Tulyakov S, Ricci E, Sebe N (2019{\natexlab{b}})
  First order motion model for image animation. In: Conference on Neural
  Information Processing Systems (NeurIPS)

\bibitem[{Song et~al.(2018)Song, Zhu, Wang, and Qi}]{song2018talking}
Song Y, Zhu J, Wang X, Qi H (2018) Talking face generation by conditional
  recurrent adversarial network. arXiv preprint arXiv:180404786

\bibitem[{Sutskever et~al.(2014)Sutskever, Vinyals, and
  Le}]{sutskever2014sequence}
Sutskever I, Vinyals O, Le QV (2014) Sequence to sequence learning with neural
  networks. In: Advances in neural information processing systems, pp
  3104--3112

\bibitem[{Suwajanakorn et~al.(2017)Suwajanakorn, Seitz, and
  Kemelmacher-Shlizerman}]{suwajanakorn2017synthesizing}
Suwajanakorn S, Seitz SM, Kemelmacher-Shlizerman I (2017) Synthesizing obama:
  learning lip sync from audio. ACM Transactions on Graphics (TOG) 36(4):1--13

\bibitem[{Szegedy et~al.(2016)Szegedy, Vanhoucke, Ioffe, Shlens, and
  Wojna}]{szegedy2016rethinking}
Szegedy C, Vanhoucke V, Ioffe S, Shlens J, Wojna Z (2016) Rethinking the
  inception architecture for computer vision. In: Proceedings of the IEEE
  conference on computer vision and pattern recognition, pp 2818--2826

\bibitem[{Thies et~al.(2016)Thies, Zollhofer, Stamminger, Theobalt, and
  Nie{\ss}ner}]{thies2016face2face}
Thies J, Zollhofer M, Stamminger M, Theobalt C, Nie{\ss}ner M (2016) Face2face:
  Real-time face capture and reenactment of rgb videos. In: Proceedings of the
  IEEE conference on computer vision and pattern recognition, pp 2387--2395

\bibitem[{Tran et~al.(2018)Tran, Yin, and Liu}]{tran2018representation}
Tran L, Yin X, Liu X (2018) Representation learning by rotating your faces.
  IEEE transactions on pattern analysis and machine intelligence
  41(12):3007--3021

\bibitem[{Vondrick et~al.(2016)Vondrick, Pirsiavash, and
  Torralba}]{vondrick2016generating}
Vondrick C, Pirsiavash H, Torralba A (2016) Generating videos with scene
  dynamics. In: Advances in neural information processing systems, pp 613--621

\bibitem[{Vougioukas et~al.(2019)Vougioukas, Petridis, and
  Pantic}]{vougioukas2019realistic}
Vougioukas K, Petridis S, Pantic M (2019) Realistic speech-driven facial
  animation with gans. International Journal of Computer Vision pp 1--16

\bibitem[{Wang et~al.(2018{\natexlab{a}})Wang, Liu, Zhu, Liu, Tao, Kautz, and
  Catanzaro}]{NIPS2018_7391}
Wang TC, Liu MY, Zhu JY, Liu G, Tao A, Kautz J, Catanzaro B
  (2018{\natexlab{a}}) Video-to-video synthesis. In: Bengio S, Wallach H,
  Larochelle H, Grauman K, Cesa-Bianchi N, Garnett R (eds) Advances in Neural
  Information Processing Systems 31, Curran Associates, Inc., pp 1144--1156

\bibitem[{Wang et~al.(2018{\natexlab{b}})Wang, Liu, Zhu, Tao, Kautz, and
  Catanzaro}]{wang2018high}
Wang TC, Liu MY, Zhu JY, Tao A, Kautz J, Catanzaro B (2018{\natexlab{b}})
  High-resolution image synthesis and semantic manipulation with conditional
  gans. In: Proceedings of the IEEE conference on computer vision and pattern
  recognition, pp 8798--8807

\bibitem[{Wang et~al.(2019)Wang, Liu, Tao, Liu, Kautz, and
  Catanzaro}]{wang2019few}
Wang TC, Liu MY, Tao A, Liu G, Kautz J, Catanzaro B (2019) Few-shot
  video-to-video synthesis. arXiv preprint arXiv:191012713

\bibitem[{Wang et~al.(2004)Wang, Bovik, Sheikh, and Simoncelli}]{wang2004image}
Wang Z, Bovik AC, Sheikh HR, Simoncelli EP (2004) Image quality assessment:
  from error visibility to structural similarity. IEEE transactions on image
  processing 13(4):600--612

\bibitem[{Wiles et~al.(2018)Wiles, Sophia~Koepke, and Zisserman}]{x2face}
Wiles O, Sophia~Koepke A, Zisserman A (2018) X2face: A network for controlling
  face generation using images, audio, and pose codes. In: Proceedings of the
  European Conference on Computer Vision (ECCV), pp 670--686

\bibitem[{Yi et~al.(2020)Yi, Ye, Zhang, Bao, and Liu}]{yi2020audio}
Yi R, Ye Z, Zhang J, Bao H, Liu YJ (2020) Audio-driven talking face video
  generation with natural head pose. arXiv preprint arXiv:200210137

\bibitem[{Yoo et~al.(2019)Yoo, Bahng, Chung, Lee, Chang, and Choo}]{yoo2019few}
Yoo S, Bahng H, Chung S, Lee J, Chang J, Choo J (2019) Coloring with limited
  data: Few-shot colorization via memory augmented networks. In: Proceedings of
  the IEEE Conference on Computer Vision and Pattern Recognition, pp
  11283--11292

\bibitem[{Zakharov et~al.(2019)Zakharov, Shysheya, Burkov, and
  Lempitsky}]{zakharov2019few}
Zakharov E, Shysheya A, Burkov E, Lempitsky V (2019) Few-shot adversarial
  learning of realistic neural talking head models. In: Proceedings of the IEEE
  International Conference on Computer Vision, pp 9459--9468

\bibitem[{Zhang et~al.(2019{\natexlab{a}})Zhang, Cheng, and
  Wang}]{zhang2019spatio}
Zhang X, Cheng F, Wang S (2019{\natexlab{a}}) Spatio-temporal fusion based
  convolutional sequence learning for lip reading. In: Proceedings of the IEEE
  International Conference on Computer Vision, pp 713--722

\bibitem[{Zhang et~al.(2019{\natexlab{b}})Zhang, Zhang, He, Li, Loy, and
  Liu}]{zhang2019one}
Zhang Y, Zhang S, He Y, Li C, Loy CC, Liu Z (2019{\natexlab{b}}) One-shot face
  reenactment. arXiv preprint arXiv:190803251

\bibitem[{Zhou et~al.(2019)Zhou, Liu, Liu, Luo, and Wang}]{zhou2019talking}
Zhou H, Liu Y, Liu Z, Luo P, Wang X (2019) Talking face generation by
  adversarially disentangled audio-visual representation. In: Proceedings of
  the AAAI Conference on Artificial Intelligence, vol~33, pp 9299--9306

\end{thebibliography}

\newpage


\begin{appendices}

\begin{figure*}[t]
  \includegraphics[width=0.98 \linewidth]{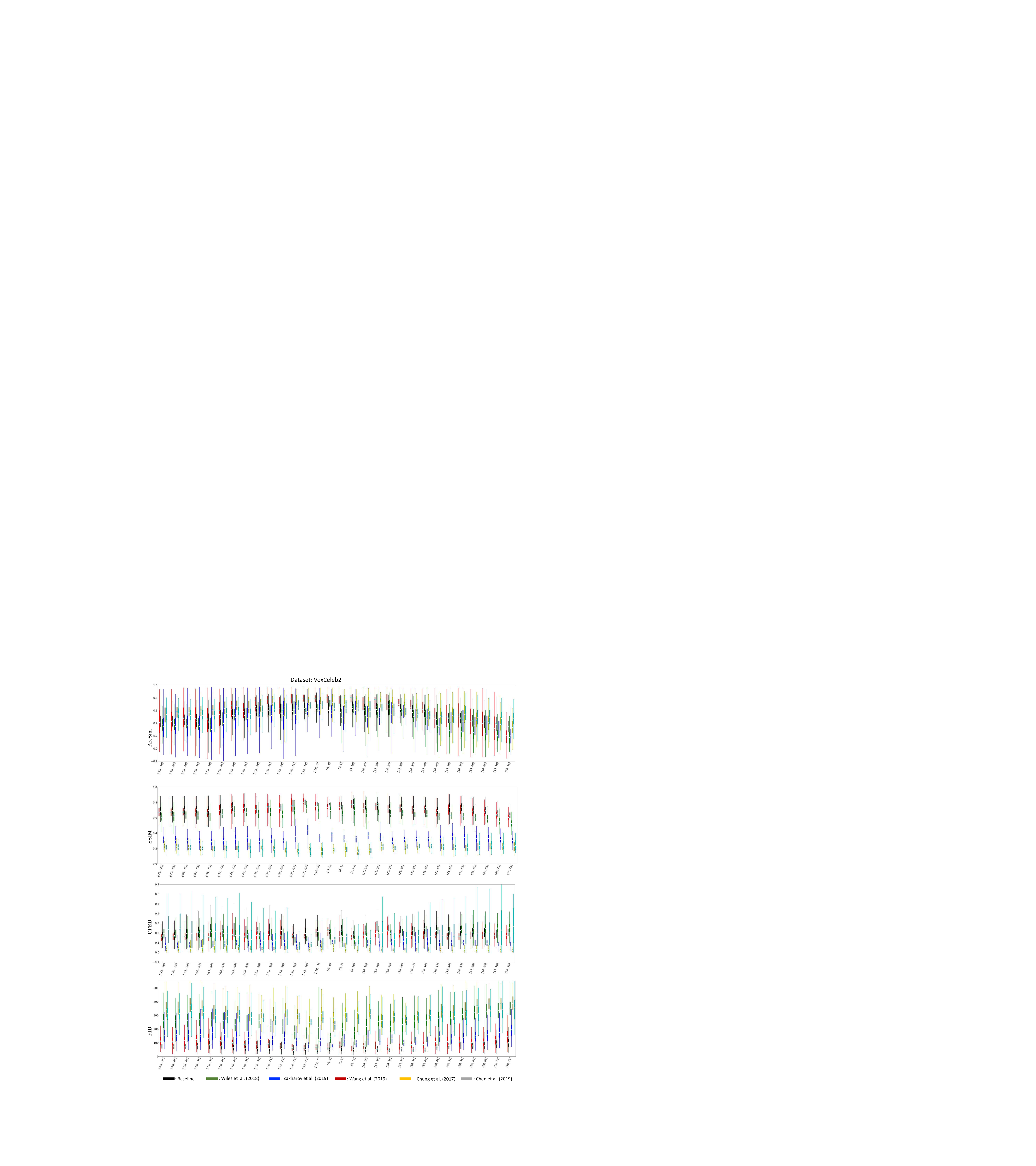}
\caption{Different evaluation metrics versus head pose degree on VoxCeleb2 dataset. Videos are divided into different bins grouped by the head pose degree. The four rows show: ArcSim, SSIM, CPBD, and FID, respectively. The X-axis is the degree bins of head pose. The last
row shows the color bars of different methods}
\label{fig:vox2_pose}       
\end{figure*}

\begin{figure*}[t]
\centering
  \includegraphics[width=0.98 \linewidth]{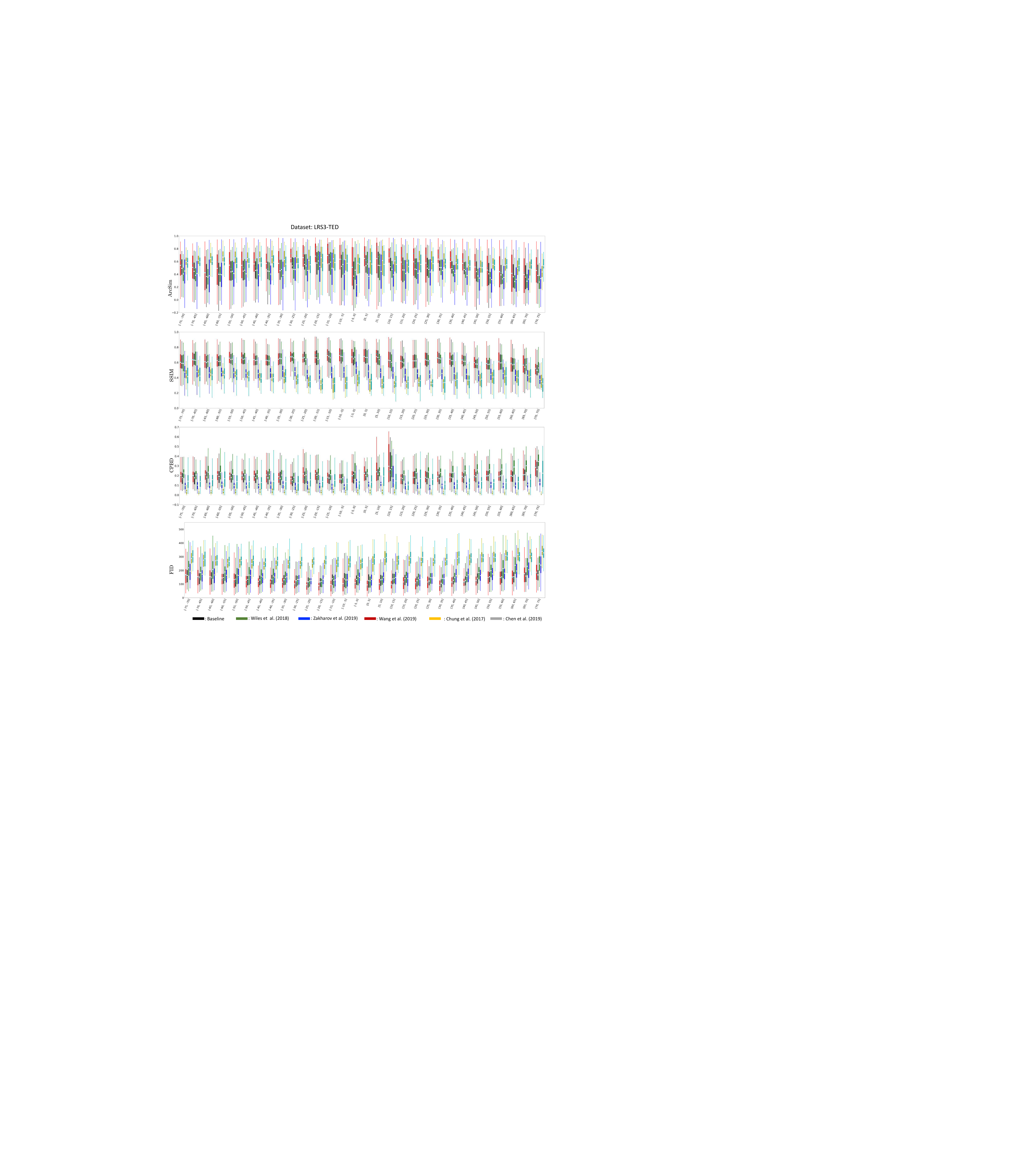}
\caption{Different evaluation metrics versus head pose degree on LRS3-TED dataset. Videos are divided into different bins grouped by the head pose degree. The four rows show: ArcSim, SSIM, CPBD, and FID, respectively. The X-axis is the degree bins of head pose. The last
row shows the color bars of different methods}
\label{fig:lrs3_pose}       
\end{figure*}

\begin{figure*}[t]
\centering
  \includegraphics[width=0.98 \linewidth]{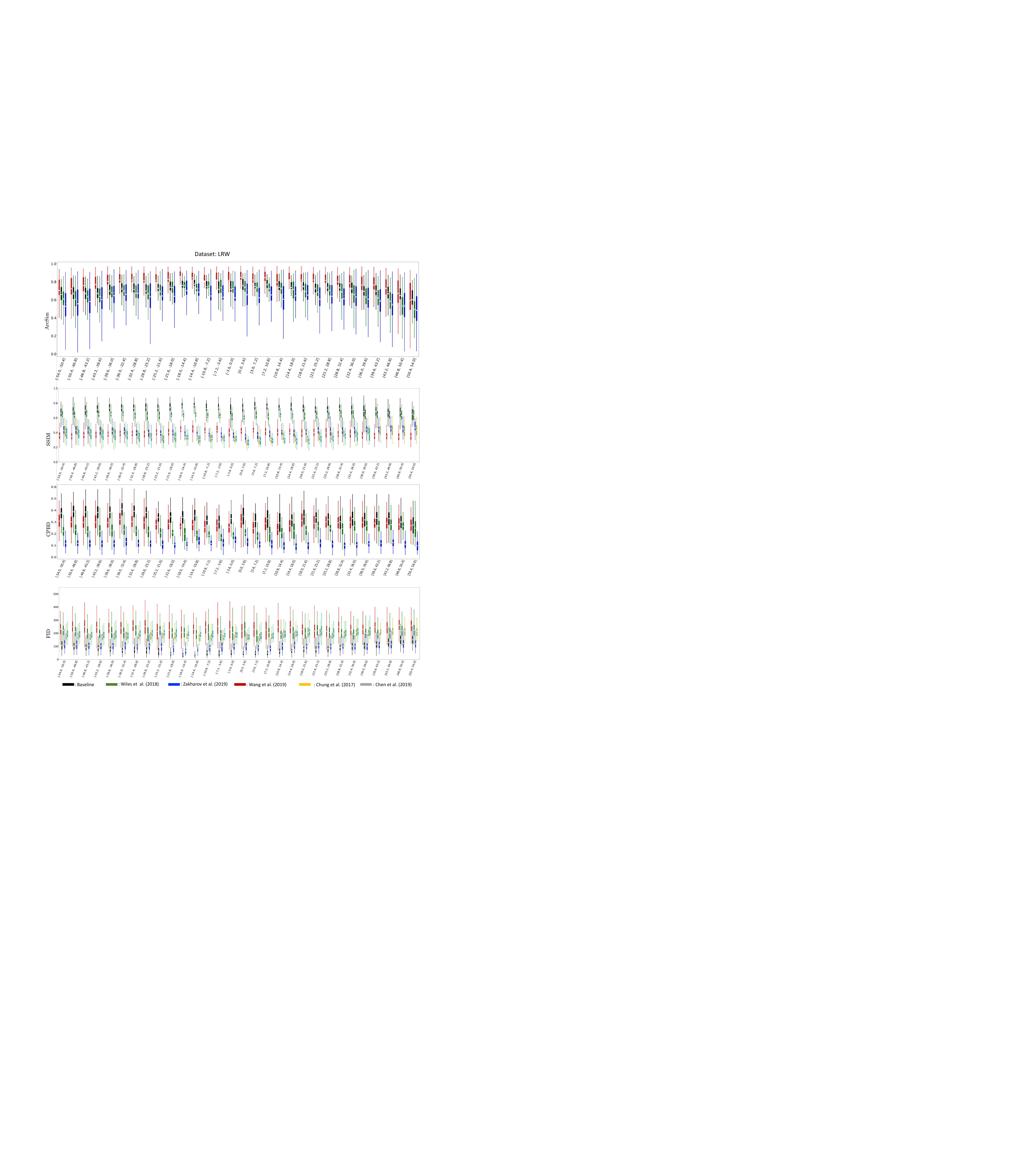}
\caption{Different evaluation metrics versus head pose degree on LRW dataset. Videos are divided into different bins grouped by the head pose degree. The four rows show: ArcSim, SSIM, CPBD, and FID, respectively. The X-axis is the degree bins of head pose. The last
row shows the color bars of different methods}
\label{fig:LRW_pose}       
\end{figure*}

We show more figures that are not included in the manuscript. The Fig.~\ref{fig:csim_all} in original manuscript shows the ArcSim scores versus the head pose on LRS3-TED and VoxCeleb2 dataset. Here, we show more evaluation metrics versus head pose on VoxCeleb2, LRS3-TED, and LRW dataset. The Fig.~\ref{fig:vox2_pose}, Fig.~\ref{fig:lrs3_pose}, and Fig.~\ref{fig:LRW_pose} show the evaluation scores versus head pose on VoxCeleb2 dataset, LRS3-TED dataset, and LRW dataset. 

\begin{figure*}[t]
\centering
  \includegraphics[width=0.78 \linewidth]{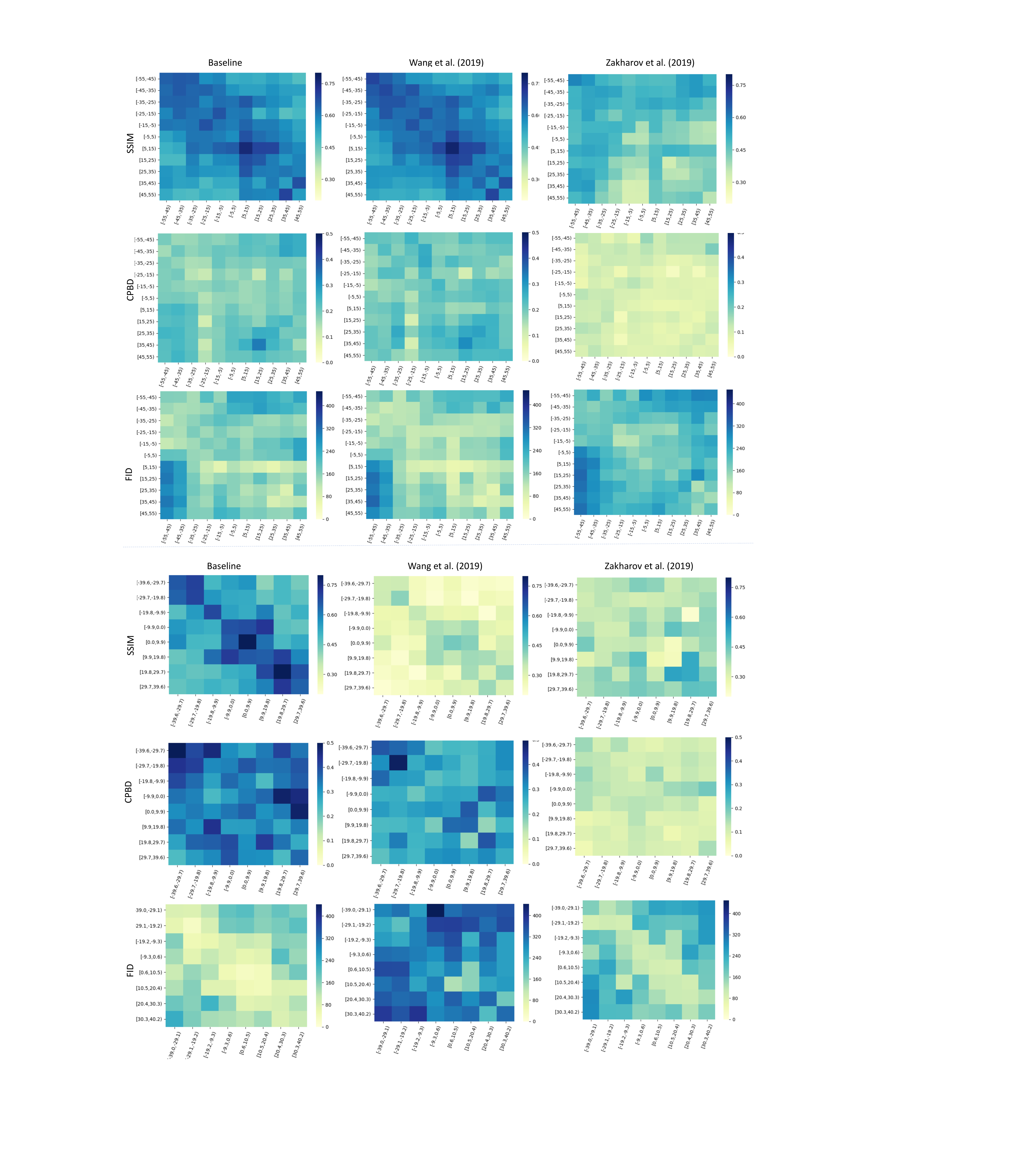}
\caption{The confusion matrix of visual quality results versus
the head motion on LRS3-TED and LRW dataset. The first three columns show the matrix on LRS3-TED dataset and the last three columns show the matrix on LRW dataset.
The X-axis and Y-axis are the head motions of reference frame
and target frame.}
\label{fig:matrix_more}       
\end{figure*}

The Fig.~\ref{fig:ssim_comfusion} in original paper shows the confusion matrix of visual quality results versus the head motion on VoxCeleb2 dataset. We also show more confusion matrix on LRS3-TED and LRW dataset in Fig.~\ref{fig:matrix_more}.

\begin{figure*}[t]
\centering
  \includegraphics[width=0.9 \linewidth]{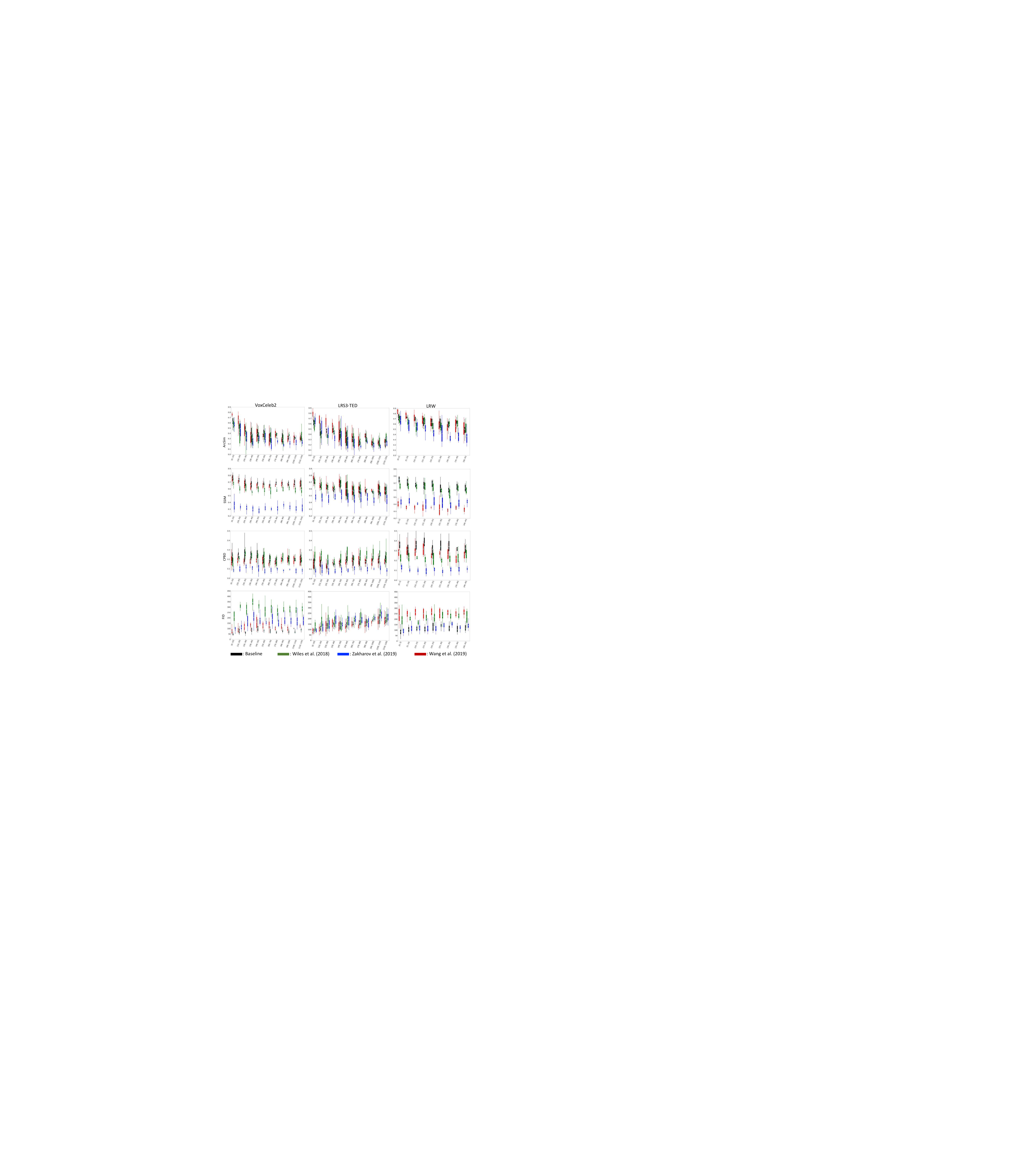}
\caption{The confusion matrix of visual quality results versus
the head motion on LRS3-TED and LRW dataset. The first three columns show the matrix on LRS3-TED dataset and the last three columns show the matrix on LRW dataset.
The X-axis and Y-axis are the head motions of reference frame
and target frame.}
\label{fig:all_motion}       
\end{figure*}

The Fig.~\ref{fig:motion_all} in original paper also shows the different evaluation metrics versus head motion score. We show more detailed comparison versus head motion score in Fig.~\ref{fig:all_motion}.

\end{appendices}

\end{document}